\documentclass[journal]{IEEEtran}
\usepackage{biblatex}
\usepackage{graphicx}
\usepackage{pdflscape}
\usepackage{lscape}
\usepackage{booktabs}
\usepackage{array}
\usepackage{color}
\usepackage{amsfonts}
\usepackage{algorithm}
\usepackage{algpseudocode}
\usepackage{amsmath}
\usepackage{multirow}
\usepackage{multicol}
\usepackage{longtable}
\usepackage{booktabs}
\usepackage[para]{threeparttable}
\usepackage{tablefootnote}
\usepackage{amssymb}
\usepackage{pifont}

\usepackage{scalerel}
\usepackage{tikz}
\usetikzlibrary{svg.path}

\definecolor{orcidlogocol}{HTML}{A6CE39}
\tikzset{
  orcidlogo/.pic={
    \fill[orcidlogocol] svg{M256,128c0,70.7-57.3,128-128,128C57.3,256,0,198.7,0,128C0,57.3,57.3,0,128,0C198.7,0,256,57.3,256,128z};
    \fill[white] svg{M86.3,186.2H70.9V79.1h15.4v48.4V186.2z}
                 svg{M108.9,79.1h41.6c39.6,0,57,28.3,57,53.6c0,27.5-21.5,53.6-56.8,53.6h-41.8V79.1z M124.3,172.4h24.5c34.9,0,42.9-26.5,42.9-39.7c0-21.5-13.7-39.7-43.7-39.7h-23.7V172.4z}
                 svg{M88.7,56.8c0,5.5-4.5,10.1-10.1,10.1c-5.6,0-10.1-4.6-10.1-10.1c0-5.6,4.5-10.1,10.1-10.1C84.2,46.7,88.7,51.3,88.7,56.8z};
  }
}\newcommand\orcidicon[1]{\href{https://orcid.org/#1}{\mbox{\scalerel*{
\begin{tikzpicture}[yscale=-1,transform shape]
\pic{orcidlogo};
\end{tikzpicture}
}{|}}}}

\usepackage[hidelinks]{hyperref}

\DeclareMathOperator*{\argmax}{\arg\!\max}

\makeatletter
\newcommand{\algmargin}{\the\ALG@thistlm}   
\makeatother
\algnewcommand{\parState}[1]{\State%
    \parbox[t]{\dimexpr\linewidth-\algmargin}{\strut #1\strut}}

\usepackage{acronym}
\acrodef{STFT}[STFT]{short-time Fourier transform}
\acrodef{RSSI}[RSSI]{radio signal strength indication}
\acrodef{VOG}[VOG]{video-oculography}
\acrodef{ERP}[ERP]{event-related potential}
\acrodef{EEG}[EEG]{electroencephalogram}
\acrodef{ECG}[ECG]{electrocardiogram}
\acrodef{FO}[FO]{face outline}
\acrodef{EDA}[EDA]{electrodermal activity}
\acrodef{BVP}[BVP]{blood volume plse}
\acrodef{PPG}[PPG]{photoplethysmography}
\acrodef{Bio-Z}[Bio-Z]{bio-impedance}
\acrodef{GYR}[GYR]{gyroscope}
\acrodef{EOG}[EOG]{electrooculography}
\acrodef{HR}[HR]{heart rate}
\acrodef{SC}[SC]{skin conductance}
\acrodef{DA}[DA]{data augmentation}
\acrodef{TL}[TL]{transfer learning}
\acrodef{MAML}[MAML]{Model-Agnostic Meta-Learning}
\acrodef{GAN}[GAN]{generative adversarial network}
\acrodef{WGAN}[WGAN]{Wasserstein generative adversarial network}
\acrodef{CGAN}[CGAN]{conditional generative adversarial network}
\acrodef{GRU}[GRU]{gated recurrent unit}
\acrodef{BERT}[BERT]{bidirectional encoder representations from transformers}
\acrodef{MLP}[MLP]{multilayer perceptron}
\acrodef{EHR}[EHR]{electronic health record}
\acrodef{IMU}[IMU]{inertial measurement unit}
\acrodef{SVM}[SVM]{support vector machine}
\acrodef{CNN}[CNN]{convolutional neural network}
\acrodef{LSTM}[LSTM]{long short-term memory}
\acrodef{GNN}[GNN]{graph neural network}
\acrodef{LBP}[LBP]{linear binary pattern}
\acrodef{EMG}[EMG]{electromyography}
\acrodef{BCI}[BCI]{brain computer interface}
\acrodef{SSVEP}[SSVEP]{steady state visually evoked potential}
\acrodef{SoC}[SoC]{system on a chip}
\acrodef{SVR}[SVR]{support vector regression}
\acrodef{GCN}[GCN]{graph convolutional network}
\acrodef{FOMAML}[FOMAML]{first-order model-agnostic meta-learning}
\acrodef{MSE}[MSE]{mean squared error}
\acrodef{LLM}[LLM]{large language model}
\acrodef{NLP}[NLP]{natural language processing}
\acrodef{ACC}[ACC]{accelerometry}
\acrodef{TEMP}[TEMP]{temperature}
\acrodef{}[]{}
\acrodef{}[]{}

\begin{document}

\title{A Survey of Few-Shot Learning for Biomedical Time Series}
\markboth{Preprint. Under review.}%
	{\MakeLowercase{\textit{et al.}}: Few-Shot Learning for Biomedical Time Series}
\author{
		Chenqi Li \orcidicon{0000-0002-0516-6988},
        Timothy Denison \orcidicon{0000-0002-5404-4004},
		Tingting Zhu \orcidicon{0000-0002-1552-5630}

        \thanks{Department of Engineering Science, University of Oxford, United Kingdom 
        
        Email: \{chenqi.li, timothy.denison, tingting.zhu\}@eng.ox.ac.uk}
}
\maketitle

\begin{abstract}
    Advancements in wearable sensor technologies and the digitization of medical records have contributed to the unprecedented ubiquity of biomedical time series data. Data-driven models have tremendous potential to assist clinical diagnosis and improve patient care by improving long-term monitoring capabilities, facilitating early disease detection and intervention, as well as promoting personalized healthcare delivery. However, accessing extensively labeled datasets to train data-hungry deep learning models encounters many barriers, such as long-tail distribution of rare diseases, cost of annotation, privacy and security concerns, data-sharing regulations, and ethical considerations. An emerging approach to overcome the scarcity of labeled data is to augment AI methods with human-like capabilities to leverage past experiences to learn new tasks with limited examples, called few-shot learning. This survey provides a comprehensive review and comparison of few-shot learning methods for biomedical time series applications. The clinical benefits and limitations of such methods are discussed in relation to traditional data-driven approaches. This paper aims to provide insights into the current landscape of few-shot learning for biomedical time series and its implications for future research and applications. 
\end{abstract}

\begin{IEEEkeywords}
    Few-Shot Learning, Biomedical Time Series
\end{IEEEkeywords}

\IEEEpeerreviewmaketitle

\section{Introduction}

Data plays a fundamental role in the success of deep learning algorithms. For biomedical applications, access to extensively labeled datasets is often a luxury due to ethical and privacy concerns associated with data acquisition and sharing. Moreover, the long-tail distribution of rare diseases leads to natural data imbalance and bias toward majority classes. Inter-subject variability further poses generalization challenges to new patients. To address these challenges, more data must be collected for the long-tail classes and individual patients to improve generalization. Alternatively, deep learning algorithms need to learn and evolve with limited labeled samples, mirroring how human adults and children can transfer past experiences to guide the learning of new tasks using a few examples. Few-shot learning emerges as a promising paradigm to augment models with capabilities to generalize effectively even when confronted with a scarcity of labeled data. The paradigm strongly resonates with meta-learning, learning how to learn, which aims to transfer knowledge from previous tasks to accelerate the learning of new tasks. 

Several surveys have provided an overview of the taxonomy of few-shot learning techniques from different perspectives \cite{jadon2023overview, song2023comprehensive, parnami2022learning, wang2021generalizing, tsoumplekas2024toward, tian2024survey}, focusing primarily on theoretical background and/or computer vision applications. For biomedical applications, existing surveys cover biomedical imaging \cite{pachetti2023systematic, kotia2021few} and text \cite{ge2022few}, and have not yet considered time series. Thus, the goal of this survey is to fill the gap and provide a literature review of few-shot learning techniques for biomedical time series applications, which play an increasingly important role in medical diagnostics, brain-computer interfaces, and wearable computing. This survey aims to answer the following research questions:
\begin{enumerate}
    \item How are few-shot learning problems defined and how do they differ from traditional deep learning pipelines? 
    \item What is the taxonomy of few-shot learning methods for biomedical time series data? 
    \item What are the applications of few-shot learning methods to biomedical time series problems and what benefits do they provide in a clinical setting? 
    \item What are the key challenges and future directions of few-shot learning for biomedical time series? 
\end{enumerate}

\section{Background and Concepts} \label{sec:background}

\begin{figure*}[!h]
    \centering
    \includegraphics[width=0.8\textwidth]{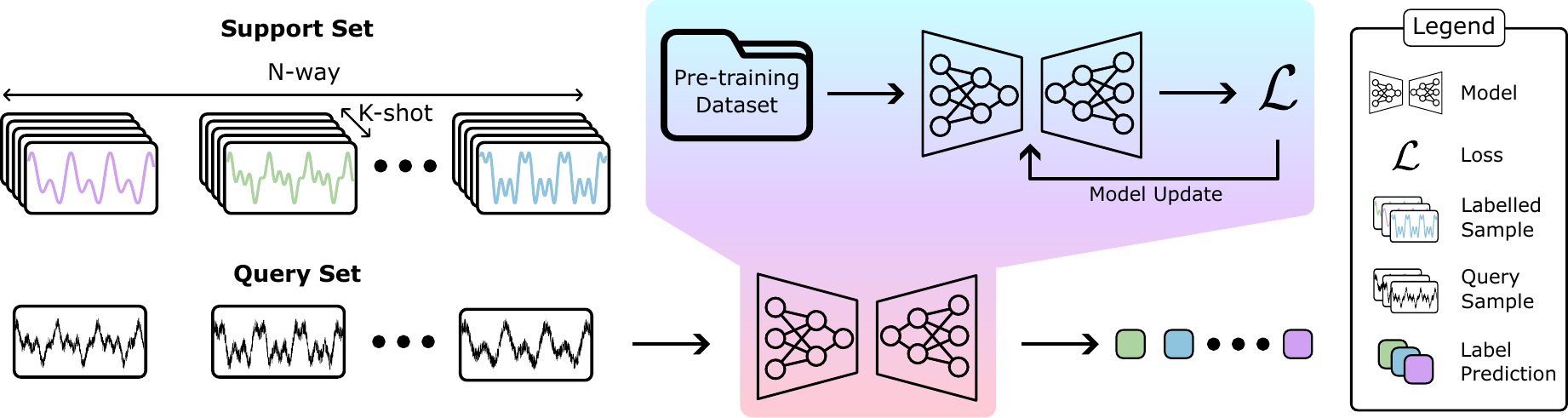}
    \caption{\textbf{Overview of few-shot learning setup.} During the pre-training stage (optional), the model is trained to learn meaningful representations and initialize its parameters to facilitate future adaption to target tasks. Subsequently, the model learns a new task using the $N$-way-$K$-shot support set and performs inference on the query set to provide label prediction.}
    \label{fig:overview}
\end{figure*}

\subsection{Problem Definition}
Unlike traditional machine learning and deep learning setups that divide the dataset into training, validation, and test subsets for different stages of the model development pipeline, few-shot learning setups are limited by the number of labeled data available and cannot afford such a division. Models trained on small datasets may memorize the specific samples instead of learning the patterns and generalizing them beyond the training set.
\par
Under the few-shot learning setting for classification, datasets are available in the form of the support set $\mathcal{S} = \{ (\mathbf{x}_{s,i}, y_{s,i}) \}_{i=1}^{N\times K}$ and the query set $\mathcal{Q} = \{ (\mathbf{x}_{q,j}, y_{q,j}) \}_{j=1}^M$ where $\mathbf{x}_{s,i}$ represents the $i$-th input of the support set of $N \times K$ samples and $\mathbf{x}_{q,j}$ represents the $j$-th input sample of the query set of $M$ samples. $y_{s,i}, y_{q,j} \in \{1,2,...,K\}$ represents the corresponding ground-truth label that belongs to one of the $K$ classes. The goal of few-shot learning is to leverage the information available from the support set $\mathcal{S}$ to estimate the label of each query sample from the query set $\mathcal{Q}$. Such setups are commonly described as $N$-way-$K$-shot problems, where $K$ indicates the number of classes and $N$ indicates the number of samples available for each class, resulting in a total of $N \times K$ samples in the support set. Figure \ref{fig:overview} provides an overview of the setup and pipeline for few-shot learning with biomedical time series. 
\par
\subsection{Episodic Pre-Training}
While the $N$-way-$K$-shot problem defines the setup of the support set, no constraint is placed on the use of different but potentially related datasets, $\mathcal{D}_{pre} = \{(\mathbf{x}_l, y_l)\}_{l=1}^{L_{pre}}$ consisting of $L_{pre}$ input-label pairs, to pre-train the few-shot learning model. The pre-trained model can then be adapted using $\mathcal{S}$ before making inferences on the query set $\mathcal{Q}$. Algorithm \ref{alg:epstrain} provides a pseudo-code for the supervised episodic pre-training paradigm. Note that during each episode of pre-training, a support set $\mathcal{S}'$ and a query set $\mathcal{Q}'$ are sampled from the pre-training dataset $\mathcal{D}_{pre}$ to mimic the few-shot learning setting. The support, query, and class size $n, m, k$ of the sampled datasets are typically set to be greater than or equal to the few-shot learning support, query, and class size $N, M, K$. For a traditional learning pipeline, the loss is computed between the ground truth label and the prediction of the model in the query sample, $\mathcal{L}(y_{q,i}, f_\theta(\mathbf{x}_{q,i}))$. For episodic pre-training, the model makes predictions using an additional support set $\mathcal{S}'$, as shown in line \ref{line:epi} of Algorithm \ref{alg:epstrain}. $\mathcal{S}'$ can be used to fine-tune the model or simply provide anchors for comparison against query samples. Episodic pre-training can also be performed in a semi-supervised manner, where the label is available for some samples and missing for others. Similarly, self-supervised episodic pre-training follows a similar procedure, but the label for each sample is generated using the dataset itself as opposed to human labeling. 
\par
\begin{algorithm}
\caption{Algorithm for Supervised Episodic Pre-Training}\label{alg:epstrain}
\begin{algorithmic}[1]
\Require $\mathcal{D}_{pre}$ pre-training dataset 
\Require $n, m, k$ support, query, class size of the sampled dataset
\Require $episodes$ number of episodes
\State Randomly initialize $\theta$
\For{$episode \gets 1$ to $episodes$}
\State Sample support set $\mathcal{S}' = \{(\mathbf{x}_{s,i}, y_{s,i})\}_{i=1}^{n\times k}$  from $\mathcal{D}_{pre}$
\State Sample query set $\mathcal{Q}' = \{ (\mathbf{x}_{q,j}, y_{q,j})\}_{j=1}^{m}$ from $\mathcal{D}_{pre}$
\State Evaluate $\mathcal{L}(y_{q,j}, f_\theta(\mathbf{x}_{q,j}, \mathcal{S}')) \indent \forall j=1,...,m$ \label{line:epi}
\State Update model parameters $\theta$ based on $\nabla_\theta \mathcal{L}$
\EndFor
\end{algorithmic}
\end{algorithm}
\par
Few-shot learning problems can be characterized based on the type of knowledge transfer a model is expected to bridge between the pre-training problem and the $N$-way-$K$-shot problem. For biomedical applications, common types of knowledge transfer include cross-session, cross-subject, cross-dataset, cross-class, cross-task, and no pre-training. In the case of cross-session few-shot learning, the model is pre-trained using data from multiple sessions of a particular subject. The model is expected to generalize to a query set from a previously unseen session of the same subject, using a $N$-way-$K$-shot support set from the same unseen session to assist prediction. Similarly, a model for cross-subject (dataset, class, task) few-shot learning is pre-trained using data from multiple subjects (datasets, classes, tasks) and is expected to generalize to unseen subjects (datasets, classes, tasks) with few samples as support. Sometimes, no pre-training is performed, the model is expected to learn with few-shot support samples from scratch and make predictions on the query set. Evidently, the type of knowledge transfer will frame few-shot learning problems of varying difficulties, and it is important to consider this difference when designing and comparing few-shot learning methods.
\par
\begin{figure*}[!h]
    \centering
    \includegraphics[width=0.8\textwidth]{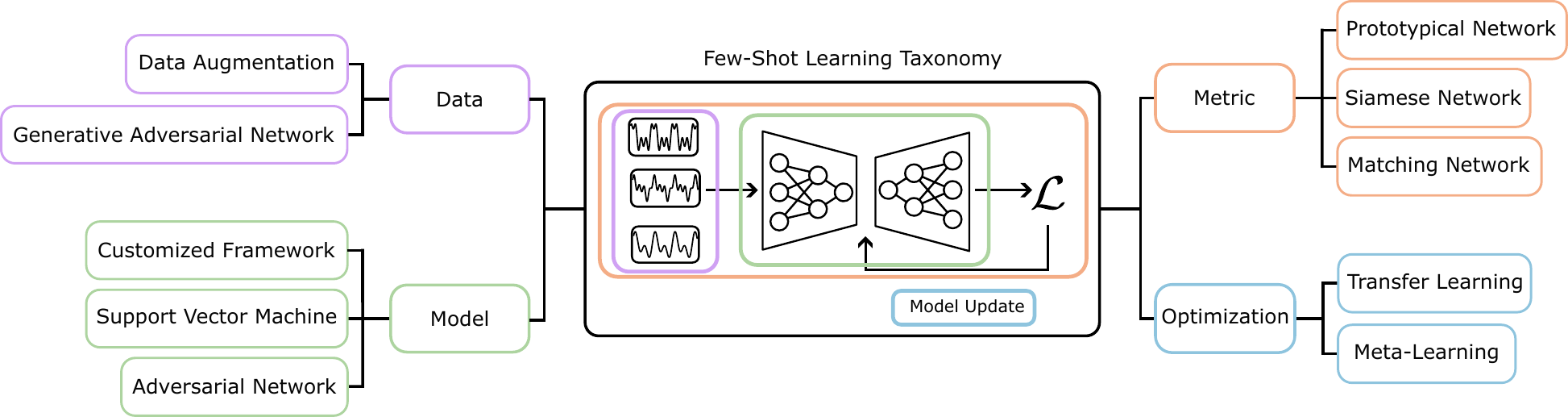}
    \caption{\textbf{Taxonomy of few-shot learning for biomedical time series.} Different few-shot learning methods propose modifications to varying parts of the training pipeline, each highlighted using a distinct color. Data-based approaches directly address the data shortage problem by generating synthetic samples to increase the size and diversity of the support set. Metric-based approaches focus on learning the similarity between samples in a representation space. Model-based approaches design model architectures to improve generalization across tasks with few samples. Optimization-based approaches guide model convergence to parameter spaces that can be quickly adjusted by fine-tuning with a few samples.}
    \label{fig:fsl_taxonomy}
\end{figure*}

To address the data shortage in the support set and facilitate the transfer of knowledge from the pre-training dataset to the $N$-way-$K$-shot problem, few-shot learning approaches propose modifications to different parts of the training pipeline to overcome the limited dataset and can be categorized into five types: data-based, model-based, metric-based, optimization-based, and hybrid methods. A taxonomy of few-shot learning is provided in Figure \ref{fig:fsl_taxonomy}. The specific methods and relevant applications will be elaborated in further detail in Section \ref{sec:results}. \par

\section{Method}
We identify and screen relevant literature using a variety of digital libraries, including ACM, PubMed, and Google Scholar. Since few-shot learning is an emerging field, especially with application to biomedical time series, no date of publication constraint was enforced. The search key was designed to capture three important characteristics of the relevant literature: "few shot learning", "biomedical", and "time series". In this work, biomedical time series is defined as time-sequential biological data of varying modalities to improve healthcare or assist clinical practices, such as diagnosis or risk prediction. The final search key uses a combination of synonyms for each characteristic:
\begin{enumerate}
    \small
    \item "shot learning"
    \item "clinical" OR "medical" OR "biomedical" OR "physiological"
    \item "EHR" OR "Electronic Health Records" OR "EEG" OR "Electroencephalogram" OR "ECG" OR "Electrocardiogram" OR "PPG" OR "Photoplethysmography" OR "EMG" OR "Electromyogram" OR "Electrooculography" OR "EOG" OR "Electrodermal Activity" OR "Actigraph" OR "Accelerometer" OR "Gyroscope" OR "IMU" OR "Inertial Measurement Unit" OR "Gait" OR "Watch" OR "Polysomnography" OR "Voice" OR "Wearables"
\end{enumerate}
ACM and PubMed returned a total of 125 and 10 articles, respectively and were all included for subsequent screening. Google Scholar returned 5,500 search results ranked in descending order of relevance. 580 articles were included for subsequent screening, after which no relevant search result has appeared over two consecutive pages. Articles from all searched databases are aggregated to remove duplicates and pre-prints. Title and abstract screening excluded records not related to few-shot learning or application to biomedical time series, but have qualified for the search criteria due to the appearance of keywords in references, future work, or background sections. Full-text screening excluded records involving biomedical images or text modalities, as existing reviews \cite{kotia2021few, ge2022few, ge2023few} have extensively covered them. A total of 55 records are included in this review.

\section{Results} \label{sec:results}

\subsection{Data-Based Few-Shot Learning}
\begin{table*}
\renewcommand{\arraystretch}{1.05}
\centering
\caption{\textbf{Data-based few-shot learning methods for biomedical time series}}
\resizebox{1\textwidth}{!}{
\begin{tabular}{ccccccccccc} 
\toprule
\toprule
 & \multicolumn{3}{c}{\textbf{Problem}} & \multicolumn{1}{c}{\textbf{Pre-Training}} & \multicolumn{1}{c}{\begin{tabular}[c]{@{}c@{}}\textbf{Knowledge} \\ \textbf{Transfer}\end{tabular}} & \multicolumn{4}{c}{\textbf{Few-Shot Learning Setup}} & \multicolumn{1}{c}{\textbf{Open Access}} \\ \cmidrule(lr){2-4} \cmidrule(lr){5-5} \cmidrule(lr){6-6} \cmidrule(lr){7-10} \cmidrule(lr){11-11}
\textbf{Reference} & \multicolumn{1}{c}{Task} & \multicolumn{1}{c}{Modality} & \multicolumn{1}{c}{Method} & \multicolumn{1}{c}{\begin{tabular}[c]{@{}c@{}}Type of \\ Pre-Training\end{tabular}} & \multicolumn{1}{c}{\begin{tabular}[c]{@{}c@{}}Type of \\ Transfer\end{tabular}} & \multicolumn{1}{c}{FSL Dataset} & \multicolumn{1}{c}{\begin{tabular}[c]{@{}c@{}}Number of \\ Classes\end{tabular}} & \multicolumn{1}{c}{Support Set Size} & \multicolumn{1}{c}{Evaluation Metric} & \multicolumn{1}{c}{\begin{tabular}[c]{@{}c@{}}Code \\ Availability\end{tabular}} \\
\midrule
\cite{you2022few} & Sleep Staging & {EEG} & {GAN} & - & N/A & SleepEDF \cite{kemp2000analysis} & 5 & 1 patient 1 night & AC, F1, CK & \ding{55} \\ \hline
 \cite{hwang2019ezsl} & Visual Stimuli & {EEG} & WGAN & Supervised & Classes & \cite{spampinato2017deep} & 10 & 0-shot & AC & \ding{55} \\ \hline
 \cite{poulain2022fewc} &\begin{tabular}[c]{@{}c@{}} Mortality\\Heart Failure\\Survival \end{tabular}& {EHR} & {GAN} & Self-Supervised & Tasks &\begin{tabular}[c]{@{}c@{}} MIMIC-IV \cite{johnson2020mimic}\\Synthea \cite{walonoski2018synthea}\\All of Us \cite{all2019all} \end{tabular}&\begin{tabular}[c]{@{}c@{}} 2\\2\\2 \end{tabular}&\begin{tabular}[c]{@{}c@{}} 1-50\%\\1-50\%\\1-50\% \end{tabular}& AUROC, AUPRC, F1 & \href{github.com/healthylaife/CEHR-GAN-BERT}{\ding{51}} \\ \hline
 \cite{suwannaphong2022radioa} & Indoor Localization &\begin{tabular}[c]{@{}c@{}} {RSSI}\\ACC \end{tabular}& DA & - & N/A & \cite{byrne2018residential} & Varying & 1-shot & F1 & \ding{55} \\ \hline
 \cite{elnaggar2023sleepa} & Sleep Posture & {IMU} & DA & - & N/A & Collected (Private) & 12 & 1-shot & AC, F1 & \ding{55} \\ \hline
 \cite{lee2023source} & Visual Stimuli & {EEG} & {GAN} & Supervised & Subjects & EEG-ImageNet40 \cite{spampinato2017deep} & 40 & \{1,2,3,4,5\}-shot & AC & \href{https://github.com/DeepBCI/Deep-BCI}{\ding{51}} \\ \hline
\bottomrule
\end{tabular}%
}
\raggedright
\begin{tablenotes}
All tasks are classification tasks. Abbreviations: Electroencephalogram (EEG), Electronic Health Record (EHR), Radio Signal Strength Indication (RSSI), Accelerometry (ACC), Inertial Measurement Unit (IMU), Accuracy (AC), Cohen's Kappa (CK), Area Under Receiver Operating Characteristic Curve (AUROC), Area Under Precision-Recall Curve (AUPRC)
\end{tablenotes}
\end{table*}

Data-based approaches aim to directly address the data shortage problem by increasing the quantity and diversity of the support set by generating synthetic samples using techniques such as \acp{GAN} or data augmentation.
\par
\subsubsection{\ac{GAN}} \label{sec:data-gan}
The core concept of \ac{GAN} \cite{goodfellow2014generative} involves adversarial training of separate discriminator and generator networks, in order to produce synthetic samples that match the distribution of real data. The goal of the generator $G$ is to mimic the distribution of the training data and create synthetic data $G(\mathbf{z})$  as a function of random noise $\mathbf{z}$ drawn from a latent space distribution $p_{\mathbf{z}}(\mathbf{z})$. Meanwhile, the discriminator $D$ aims to distinguish between training data $\mathbf{x}$ drawn from the training data distribution $p_{d}(\mathbf{x})$ and synthetically generated data $G(\mathbf{z})$. The discriminator learns to estimate the probability that the given input is real or synthetic. Networks are backpropagated together using an adversarial loss function $V(D, G)$, in which the discriminator aims to maximize while the generator aims to minimize:
\begin{equation*}
\resizebox{\columnwidth}{!}{
$\displaystyle
\min_{G}\max_{D}V(D, G) = \mathop{\mathbb{E}}_{\mathbf{x} \sim p_d(\mathbf{x})}[\log D(\mathbf{x})] + \mathop{\mathbb{E}}_{\mathbf{z} \sim p_{\text{z}}(\mathbf{z})}[\log(1 - D(G(\mathbf{z})))]
$
}
\end{equation*}
\par
Several variants of \ac{GAN} have since been proposed to address the challenges associated with the training and generation process of \ac{GAN}. Notably, mode collapse and vanishing gradients are widely recognized training challenges for \ac{GAN} \cite{arjovsky2017towards}. To address these challenges and improve training stability, \acp{WGAN} propose to use Wasserstein distance as a training objective to measure the similarity between the real and synthetic data distributions \cite{arjovsky2017wasserstein}. On another note, \ac{CGAN} pass a variable or label $y$ as an additional input to the discriminator and the generator, in order to control the generation process and synthesize data that conform to specific conditions or constraints \cite{mirza2014conditional}.
\par
Many works have focused on using \acp{GAN} to diversify and expand the size of the few-shot support set, in hopes of using the expanded support set to train a better performing model. For example, You et al. \cite{you2022few} propose SleepGAN which consists of a \ac{WGAN} and a relational memory generator to generate synthetic \ac{EEG} sleep epochs and transition sequences from the sleep stage. To accelerate training and generation performance, the authors adopted the generator and discriminator architectures of ConSinGAN \cite{hinz2021improved}. To demonstrate the effectiveness of SleepGAN in generating high-quality samples from few-shot real data, a downstream sleep staging model was trained using a combination of the patient's few-shot real data and synthetically generated epochs. An accuracy of 81. 1\% was observed, compared to 77. 5\% when training the same sleep staging model using the patient's few-shot real data alone. Similarly, Lee et al. \cite{lee2023source} generate synthetic \ac{EEG} samples for the problem of source-free subject adaptation. Source-free subject adaptation refers to the transfer of knowledge from source subjects to target subjects, using few-shot target subject data and a pre-trained model on the source subject data, without access to any source subject data itself. The authors employ a modified version of \ac{GAN}. The discriminator is a frozen network pre-trained on the source subject data, and it supervises the generator to synthesize source subject samples that align with the source distribution. The generated source samples can then be combined with few-shot target samples for subject-independent feature learning. Without relying on real source subject data, few-shot subject transfer is achieved while protecting patient privacy and bypassing barriers to data sharing.
\par
Learning to synthesize model representations, as opposed to raw time series, can also be considered a form of data augmentation to supplement the few-shot support set. Hwang et al. \cite{hwang2019ezsl} propose a three-stage framework for zero-shot learning of visual stimuli classification, which aims to predict the image class presented to the subject based on the given \ac{EEG} response. First, a \ac{GRU}-based encoder is trained to extract features from real \ac{EEG} signals. Second, a \ac{WGAN} conditioned on word2vec \cite{mikolov2013efficient} vectors of a given semantic class is trained to generate synthetic features that mimic the real features extracted by the \ac{GRU} from real \ac{EEG} signals. Thirdly, the word2vec semantic vector of each unseen class is provided as input to the \ac{WGAN}, and the generated features are used to train a classifier for zero-shot classification, achieving 39.65\% zero-shot ten-class accuracy. The pipeline relies heavily on the assumption that \ac{WGAN} is capable of generalizing and generating \ac{GRU} feature vectors for unseen classes. Similarly, Poulain et al. \cite{poulain2022fewc} propose to simulate representations of \ac{BERT} outputs using a \ac{MLP}-based \ac{GAN}. The generator learns to simulate synthetic representation, while the discriminator learns to perform two tasks: differentiating between real and synthetic representations through an unsupervised loss and producing classification predictions through a supervised loss. Therefore, the discriminator not only provides supervision to the generator but also acts as the classifier for downstream tasks. In addition to the few-shot labeled dataset, the authors further leverage an unlabelled dataset to provide a more comprehensive representation of the real samples for comparison against synthetic samples. Experimental results on multiple \ac{EHR} predictive tasks show that the proposed network surpasses state-of-the-art \ac{EHR} predictive models, especially under the few-shot setting. However, the assumption that an unlabeled dataset is available is often invalid in the few-shot learning setting, limiting the generalization of this approach to other tasks and datasets. 
\par
\subsubsection{Data Augmentation} \label{sec:dataaugmentation}
Data augmentation for time series can be divided into time, frequency, and time-frequency domain-based methods. Time domain methods manipulate time series directly, such as noise injection, flipping, cropping, scaling, imputation, amplitude scaling, and up/downsampling. Frequency domain methods apply the Fourier transform to convert signals from the time domain to the frequency domain, perturbing amplitude and frequency representations to introduce variations. Time-frequency domain methods introduce variations to time-frequency representations obtained through a variety of transformations, such as wavelet transform, \ac{STFT}, and spectrogram extraction.
\par
Suwannaphong et al. \cite{suwannaphong2022radioa} studied two augmentation techniques for the localization of indoor patients using \ac{RSSI} sensors, namely, adding Gaussian noise and dropping out channels. The results show that the augmentation techniques are simple but effective in boosting the one-shot classification with random forest, reducing the data collection time from hours to minutes. In a different application, Elnaggar et al. \cite{elnaggar2023sleepa} investigated data augmentation for \ac{IMU} sensor. Due to the mathematical constraints of the quaternion representation, simply adding noise to the quaternion creates non-sensible synthetic data. To overcome this challenge,  the quaternion representations are converted to axis-angle representations, and Gaussian noise is injected into axis and angle components to generate near-realistic postural data. The classification of sleep posture is subsequently performed through an ensemble of soft margin \ac{SVM}. Experimental results show that noise augmentation helps to improve the robustness of the classifier against intra-posture similarity, given the appropriate degree of noise level. It should be noted that both data augmentation-based approaches from Section \ref{sec:dataaugmentation} are applied to traditional machine learning models that typically require less data than deep learning models. Data augmentation is likely insufficient as a standalone method to help overcome the data shortage barrier in few-shot learning settings for deep learning models.

\par
\subsubsection{Summary} \label{sec:datasummary}
Data-based methods excel at generating realistic synthetic samples and feature representations to expand the few-shot support set, contributing to performance gains in a variety of downstream tasks. However, a major concern for data-based methods is that the support set fails to capture a representative distribution of the data. Thus, the distribution of the augmented set will be similar to the distributions of the un-augmented set, potentially leading to overfitting on the skewed distribution of the support set. 

\par

\subsection{Metric-Based Few-Shot Learning}

\begin{table*}
\renewcommand{\arraystretch}{1.05}
\centering
\caption{\textbf{Metric-based few-shot learning methods for biomedical time series}}
\resizebox{1\textwidth}{!}{
\begin{tabular}{ccccccccccc} 
\toprule
\toprule
 & \multicolumn{3}{c}{\textbf{Problem}} & \multicolumn{1}{c}{\textbf{Pre-Training}} & \multicolumn{1}{c}{\begin{tabular}[c]{@{}c@{}}\textbf{Knowledge} \\ \textbf{Transfer}\end{tabular}} & \multicolumn{4}{c}{\textbf{Few-Shot Learning Setup}} & \multicolumn{1}{c}{\textbf{Open Access}} \\ \cmidrule(lr){2-4} \cmidrule(lr){5-5} \cmidrule(lr){6-6} \cmidrule(lr){7-10} \cmidrule(lr){11-11}
\textbf{Reference} & \multicolumn{1}{c}{Task} & \multicolumn{1}{c}{Modality} & \multicolumn{1}{c}{Method} & \multicolumn{1}{c}{\begin{tabular}[c]{@{}c@{}}Type of \\ Pre-Training\end{tabular}} & \multicolumn{1}{c}{\begin{tabular}[c]{@{}c@{}}Type of \\ Transfer\end{tabular}} & \multicolumn{1}{c}{FSL Dataset} & \multicolumn{1}{c}{\begin{tabular}[c]{@{}c@{}}Number of \\ Classes\end{tabular}} & \multicolumn{1}{c}{Support Set Size} & \multicolumn{1}{c}{Evaluation Metric} & \multicolumn{1}{c}{\begin{tabular}[c]{@{}c@{}}Code \\ Availability\end{tabular}} \\
\midrule
\cite{wang2023generalized} & {SSVEP} & {EEG} & Matching & Supervised & Classes &\begin{tabular}[c]{@{}c@{}} BETA \cite{liu2020beta}\\Benchmark \cite{wang2016benchmark} \end{tabular}&\begin{tabular}[c]{@{}c@{}} 8\\8 \end{tabular}&\begin{tabular}[c]{@{}c@{}} 0-shot\\0-shot \end{tabular}& AC & \ding{55} \\ \hline
 \cite{compagnon2020learninga} & Human Activity & {IMU} & Matching & Self-Supervised & Datasets & MobiActV2 \cite{chatzaki2017human} & 12 & 1-shot & AC & \ding{55} \\ \hline
 \cite{sun2023few} &\begin{tabular}[c]{@{}c@{}} Arrhythmia\\Face Outline\\Hand Gesture \end{tabular}&\begin{tabular}[c]{@{}c@{}} ECG\\FO\\ACC \end{tabular}& Prototypical & - & N/A &\begin{tabular}[c]{@{}c@{}} MIT-BIH \cite{goldberger2000physiobank}\\UCR FaceAll \cite{dau2019ucr}\\UCR UWave \cite{dau2019ucr} \end{tabular}&\begin{tabular}[c]{@{}c@{}} 9\\14\\24 \end{tabular}&\begin{tabular}[c]{@{}c@{}} 5-shot\\5-shot\\5-shot \end{tabular}& AC & \ding{55} \\ \hline
 \cite{hernandez2023prototypical} & Speech Imagery & {EEG} & Prototypical & Supervised, Epiodic & Subjects &\begin{tabular}[c]{@{}c@{}} KaraOne \cite{zhao2015classifying}\\ASU \cite{nguyen2017inferring} \end{tabular}&\begin{tabular}[c]{@{}c@{}} 2, 11\\2, 8 \end{tabular}&\begin{tabular}[c]{@{}c@{}} \{3,5,7,10\}-shot\\\{3,5,7,10\}-shot \end{tabular}& AC & \ding{55} \\ \hline
 \cite{ahn2023automatic} & Stridor & Audio & Prototypical & Supervised & Tasks & Collected (Private) & 2 & \{4,6,8\}-shot & AC, SP, SE, PR, F1 & \href{https://github.com/kskim-phd/PFL-SD}{\ding{51}} \\ \hline
 \cite{tang2020interpretable} &\begin{tabular}[c]{@{}c@{}} Myocardial Infarction\\Arrhythmia \end{tabular}& {ECG} & Prototypical & - & N/A &\begin{tabular}[c]{@{}c@{}} UCR ECG200 \cite{dau2019ucr}\\UCR TwoLeadECG \cite{dau2019ucr}) \end{tabular}&\begin{tabular}[c]{@{}c@{}} 2\\2 \end{tabular}&\begin{tabular}[c]{@{}c@{}} \{6,8\}-shot\\\{6,8\}-shot \end{tabular}& AC & \href{https://github.com/Wensi-Tang/DPSN}{\ding{51}} \\ \hline
 \cite{pan2023multiplea} & Consciousness & {EEG} & Prototypical & Supervised, Episodic & Sessions & Collected (Private) & 2 & \{5,10,15\}-shot & AC, RE, PR, F1 & \ding{55} \\ \hline
 \cite{salekin2021understanding} & {ASD} & {EEG} & Prototypical & Supervised, Episodic & Subjects & Collected (Private) & {4, 8} & 3-shot & AC & \ding{55} \\ \hline
 \cite{suo2020tadanet} & Disease & {EHR} & Prototypical & Supervised, Episodic & Classes & MIMIC-III \cite{johnson2016mimic} & {3, 5} & \{1,5\}-shot & AC & \ding{55} \\ \hline
 \cite{li2021oneb} & Arrhythmia & {ECG} & Siamese & - & N/A & MIT-BIH \cite{goldberger2000physiobank} & 5 & \{1,5,10,30,50\}-shot & AC & \ding{55} \\ \hline
 \cite{zhang2022few} & Emotion &\begin{tabular}[c]{@{}c@{}} {EDA}\\{BVP}\\TEMP \end{tabular}& Siamese & - & N/A &\begin{tabular}[c]{@{}c@{}} CASE \cite{sharma2019dataset}\\RCEA \cite{zhang2020rcea}\\CEAP-360VR \cite{xue2021ceap} \end{tabular}&\begin{tabular}[c]{@{}c@{}} {2, 5}\\{2, 5}\\{2, 5} \end{tabular}&\begin{tabular}[c]{@{}c@{}} \{1,5,10\}-shot\\\{1,5,10\}-shot\\\{1,5,10\}-shot \end{tabular}& AC, Macro-F1 & \href{https://github.com/cwi-dis/EmoDSN}{\ding{51}} \\ \hline
 \cite{munia2021imbalanceda} & Seizure & {EEG} & Siamese & Supervised & Sessions & CHB-MIT \cite{shoeb2009application} & 2 & 1-shot & SP, SE, AC, F1, PR & \href{https://github.com/cwi-dis/EmoDSN}{\ding{51}} \\ \hline
 \cite{wang2023similarity} & Hand Gesture & {EMG} & Siamese & Supervised & Classes & Collected (Private) & 5 & \{1,5\}-shot & AC & \ding{55} \\ \hline
 \cite{gupta2021similarityb} & Arrhythmia & {ECG} & Siamese & Supervised & Datasets & MIT-BIH \cite{goldberger2000physiobank} & 5 & 5-shot & AC & \ding{55} \\ \hline
 \cite{soroushmojdehi2022transfer} & Hand Gesture & {EMG} & Siamese & Supervised & Classes & Collected \cite{soroushmojdehi2022transfer} & 6 & 5-shot & AC & \href{github.com/RahilSoroush/Nearlab_sEMG_dataset}{\ding{55}} \\ \hline
 \cite{tam2022siamese} & Hand Gesture & {EMG} & Siamese & Supervised & Sessions & Collected (Private) & 6 & \{0-20\}-shots & AC & \ding{55} \\ \hline
 \cite{ng2023few} & Atrial Fibrillation & {ECG} & Siamese & Supervised & Subjects &\begin{tabular}[c]{@{}c@{}} MIT-BIH \cite{goldberger2000physiobank}\\LT-AF \cite{petrutiu2007abrupt} \end{tabular}&\begin{tabular}[c]{@{}c@{}} 2\\2 \end{tabular}&\begin{tabular}[c]{@{}c@{}} \{1,3,5,7,9,11\}-shot\\\{1,3,5,7,9,11\}-shot \end{tabular}& Macro-F1, AC, SE, SP & \ding{55} \\ \hline
 \cite{moin2018emg} & Hand Gesture & {EMG} & Prototypical & - & N/A & Collected \cite{moin2018emg} & 5 & \{1-10\}-shot & AC & \href{https://github.com/a-moin/flexemg}{\ding{51}} \\ \hline
 \cite{bhosale2022calibration} & Emotion & {EEG} & Matching & Supervised & Subjects & DEAP \cite{koelstra2011deap} & 2 & \{5,15,20,25\}-shot & AC & \ding{55} \\ \hline
 \cite{mccartney2019zeroa} & Visual Stimuli & {EEG} & Matching & Supervised & Classes &\begin{tabular}[c]{@{}c@{}} Trento \cite{murphy2009eeg}\\\cite{kaneshiro2015representational} \end{tabular}&\begin{tabular}[c]{@{}c@{}} 60\\6 \end{tabular}&\begin{tabular}[c]{@{}c@{}} 0-shot\\0-shot \end{tabular}& CMC, AUROC & \ding{55} \\ \hline
 \cite{burrello2019hyperdimensionala} & Seizure & {EEG} & Prototypical & - & N/A & Collected \cite{burrello2019hyperdimensionala} & 2 & \{2-14\}-shot & SP, SE, AC & \ding{55} \\ \hline
\bottomrule
\end{tabular}%
}
\raggedright
\begin{tablenotes}
All tasks are classification tasks. Abbreviations: Electroencephalogram (EEG), Inertial Measurement Unit (IMU), Electrocardiogram (ECG), Face Outline (FO), Accelerometry (ACC), Electronic Health Record (EHR), Electrodermal Activity (EDA), Blood Volume Pulse (BVP), Temperature (TEMP), Electromyography (EMG), Accuracy (AC), Specificity (SP), Sensitivity (SE), Precision (PR), Recall (RE), Cumulative Match Curve (CMC), Area Under Receiver Operating Characteristic Curve (AUROC)
\end{tablenotes}
\end{table*}
Metric-based methods focus on learning the similarity or distance between data points through an embedding network. The embedding network projects data points to a representation space where instances of similar labels are clustered closely together, while those of different labels are positioned farther apart.
\subsubsection{Siamese Network}
Siamese network consists of parallel weight-sharing embedding networks and similarity functions that project and measure similarity between paired input data points in the representation space \cite{bromley1993signature}. Given a $N$-way-$K$-shot support set $\mathcal{S} = \{ (\mathbf{x}_{s,i}, y_{s,i}) \}_{i=1}^{N\times K}$ and a query set $\mathcal{Q} = \{ (\mathbf{x}_{q,j}, y_{q,j}) \}_{j=1}^M$, Siamese networks compute the similarity between the query and support samples using a similarity metric consisting of a distance function and a sigmoid activation. The similarity between the $j$-th query sample $\mathbf{x}_{q,j}$ and the $i$-th support sample $\mathbf{x}_{s,i}$ can be computed as:
\begin{equation*}
s(\mathbf{x}_{q,j}, \mathbf{x}_{s,i}) = \frac{1}{1+\text{exp}(-||f_\theta(\mathbf{x}_{q,j}), f_\theta(\mathbf{x}_{s,i})||)}
\end{equation*}
where $||f_\theta(\mathbf{\mathbf{x}}_{q,j}), f_\theta(\mathbf{\mathbf{x}}_{s,i})||$ is the distance function (e.g. Euclidean distance, cosine similarity) between embeddings of paired inputs to the Siamese network and $f_\theta$ represents the embedding network parameterized by parameters $\theta$. The classification prediction $\hat y_{q,j}$ for query sample $\mathbf{x}_{q,j}$ is defined as:
\begin{equation*}
\hat y_{q,j}  = y_{s, \hat i} \indent \text{where} \indent \hat i = \argmax_{i} s(\mathbf{x}_{q,j}, \mathbf{x}_{s,i})
\end{equation*}
The choice of parallel weight-sharing embedding network and similarity metric is highly flexible and dependent on the application and modality. Common choices of embedding networks include \ac{CNN} \cite{bertinetto2016fully} \& transformer \cite{bandara2022transformer} for images \& time series, \ac{LSTM} \cite{mueller2016siamese} for text, and \ac{GNN} \cite{jin2021multi} for graphs. Siamese networks can be trained using the cross-entropy loss, but contrastive and triplet loss that are designed for learning effective and discriminative embeddings are also widely used to train Siamese networks. Given an embedding network $f_\theta$ parameterized by $\theta$, the contrastive loss between the $i$-th support pair $(\mathbf{x}_{s,i}, y_{s,i})$ and the $j$-th query pair $(\mathbf{x}_{q,j}, y_{q,j})$ can be defined as:
\begin{align*}\resizebox{0.9\hsize}{!}{$
\mathcal{L}_c = (1-y) \cdot ||f_\theta{(\mathbf{x}_{s,i})},f_\theta{(\mathbf{x}_{q,j})}||^2 + y \cdot \max{(0, m-||f_\theta{(\mathbf{x}_{s,i})},f_\theta{(\mathbf{x}_{q,j})}||^2)}$
}
\end{align*}
where $y$ is a binary label for whether input pairs are similar $y=0$ if $y_{s,i}=y_{q,j}$ or dissimilar $y=1$ if $y_{s,i} \neq y_{q,j}$, and $m>0$ is a threshold that will not push dissimilar pairs further apart if the distance between the dissimilar pairs is sufficiently large. The triplet loss further extends the contrastive loss to include comparison against both similar and dissimilar samples simultaneously. Given the $j$-th query pair $(\mathbf{x}_{q,j}, y_{q,j})$ as an anchor, the $i^+$-th support pair $(\mathbf{x}_{s,i^+}, y_{s,i^+})$ as a similar example to the anchor where $y_{q,j} = y_{s,i^+}$, and the $i^-$-th support pair $(\mathbf{x}_{s,i^-}, y_{s,i^-})$ as a dissimilar example to the anchor where $y_{q,j} \neq y_{s,i^-}$, the triplet loss is defined as:
\begin{align*} \resizebox{0.9\hsize}{!}{$
\mathcal{L}_t = \max{(0, ||f_\theta{(\mathbf{x}_{q,j})}, f_\theta{(\mathbf{x}_{s,i^+})}|| - ||f_\theta{(\mathbf{x}_{q,j})}, f_\theta{(\mathbf{x}_{s,i^-})}|| + \alpha)}$
}\end{align*}
where $\alpha$ is a margin that enforces a minimum difference between the embedding distance from the anchor to the positive sample and from the anchor to the negative sample, to avoid convergence to the trivial solution of projecting all samples to the same point in the embedding space.
\par
The effectiveness of Siamese \ac{CNN} for few-shot learning has been verified across many biomedical time series applications of varying modalities. Major differences across Siamese \ac{CNN} lie in the choice of embedding networks and similarity metrics. 1D CNN-based Siamese networks are by far the most common choice among applications to biomedical time series. Gupta et al. \cite{gupta2021similarityb} calculated similarity based on sigmoid activation of L1 distance between feature pairs from 1D \ac{CNN} embedding network and show superior few-shot learning \ac{ECG} classification performance than dynamic time warping and \ac{LSTM}-based methods. Soroushmojdehi et al. \cite{soroushmojdehi2022transfer} adopted a similar setup but added a fully connected layer, fine-tuned with the support set, between the embedding network and the L1 distance function. On the other hand, Li et al. \cite{li2021oneb} chose to add a fully connected layer between the L1 distance function and the sigmoid activation. Ng et al. \cite{ng2023few} also adopted a similar architecture as \cite{gupta2021similarityb} and replaced the sigmoid activation function with a two-layer fully connected network with two output neurons for direct binary classification of input pairs as similar or dissimilar. Classification with Siamese networks relies heavily on the assumption that the labels for each data point are accurate and often do not hold in emotion recognition datasets that involve fine-grained, self-reported labels. To overcome this problem, Zhang et al. \cite{zhang2022few} propose to replace the averaging of similarity score for each class with a Bayesian fusion-based distance module to minimize overfitting to mislabelled data points. All the aforementioned Siamese 1D \ac{CNN} methods demonstrate improvements in few-shot learning performance compared to existing backbones trained without the Siamese framework. 
\par
Many works have also sought to convert raw time series to alternative representations. For example, Munia et al. \cite{munia2021imbalanceda} propose to symbolize \ac{EEG} signal as 1D \ac{LBP} histogram representation, in an attempt to capture wave morphology of local neighborhoods and summarize structural patterns of the signal. The authors further trained a Siamese network with 1D \ac{CNN} embedding network and sigmoid-activated fully connected layer as similarity metric. The results show improvements over baseline \ac{CNN} under the few-shot learning setting but do not show consistent improvements when trained using all available data. In another work, Tam et al. \cite{tam2022siamese} map individual time series segments of each \ac{EMG} channel into a 2D muscle activity heatmap based on the physical layout of the channels. A personalized Siamese network with 2D \ac{CNN} embedding network and cosine similarity metric demonstrate strong few-shot learning transfer across user sessions, with near-perfect performance for at least half of the participants. Similarly, Wang et al. \cite{wang2023similarity} map time series segments into four 2D muscle activity heatmaps, arranged by the physical layout of the channels. Each heatmap corresponds to one of the four time-domain features originally proposed by \cite{hudgins1993new}: mean absolute value, number of zero-crossings, number of slope sign changes, and waveform length. Siamese network based on 2D \ac{CNN} embedding network and sigmoid-activated Euclidean distance function was found to demonstrate strong generalization to new classes and subjects for hand gesture recognition under the few-shot learning setting.

\subsubsection{Matching Network}
Matching networks \cite{vinyals2016matching} follow a similar procedure as Siamese networks, but instead of comparing pairs of query and support samples, the query sample is compared against the entire support set. Furthermore, the embedding network for support and query samples may not necessarily share the same weights or architecture. Given a $N$-way-$K$-shot support set $\mathcal{S} = \{ (\mathbf{x}_{s,i}, y_{s,i}) \}_{i=1}^{N\times K}$ and query set $\mathcal{Q} = \{ (\mathbf{x}_{q,j}, y_{q,j}) \}_{j=1}^M$, matching networks compute the similarity between the support and query sample using a similarity metric consisting of a distance function and a softmax activation. The similarity between the $j$-th query sample $\mathbf{x}_{q,j}$ and the $i$-th support sample $\mathbf{x}_{s,i}$ is:
\begin{equation*}
    s(\mathbf{x}_{q,j}, \mathbf{x}_{s,i}) = \frac{\text{exp}(-||f_\theta(\mathbf{x}_{q,j}), g_\phi(\mathbf{x}_{s,i})||)}{\sum\limits_{i'=1}^{N\times K}\text{exp}(-||f_\theta(\mathbf{x}_{q,j}), g_\phi{(\mathbf{x}_{s,i'})}||)}
\end{equation*}
where $f_\theta$ and $g_\phi$ are embedding networks for query and support samples parameterized by $\theta$ and $\phi$ respectively. The final prediction $\hat y_{q,j}$ for query sample $x_{q,j}$ then becomes a linear combination of labels weighted by similarity score:
\begin{equation*}
    \hat y_{q,j} = \sum_{i=1}^{N\times K}{s(\mathbf{x}_{q,j}, \mathbf{x}_{s,i}) \cdot y_{s,i}}
\end{equation*}
The similarity score can be interpreted as an attention mechanism that weighs the label of each support sample.
\par
One of the first applications of the matching network to biomedical time series was performed by Compagnon et al. \cite{compagnon2020learninga} using a \ac{GRU}-based embedding network and a softmax activated Euclidean or cosine distance function. The embedding network consists of a \ac{GRU} encoder to produce embeddings for the support and query samples, which are further strengthened by a bidirectional \ac{GRU} and an attention \ac{GRU} to produce context embeddings for measuring similarity. The best performance was observed when the encoder network was sequence-to-sequence pre-trained on a similar dataset, which achieved a comparable result to baseline networks trained on the entirety of the target dataset. Similarly, Bhosale et al. \cite{bhosale2022calibration} also employed Euclidean distance to match the query sample to the support set, but the embedding network consists of 3D convolution encoder and bi-directional \ac{LSTM} to learn temporal relations. An episodic pre-training paradigm was adopted and different sampling techniques for forming the support and query sets were explored. Random sampling draws samples randomly without any constraint and is the most widely used sampling technique for episodic pre-training. Subject-dependent sampling draws support samples that belong to the same subject as the query samples, while subject-independent sampling draws support samples that belong to different subjects as the query samples. Quantitative results show that the choice of sampling strategy does indeed affect few-shot learning performance. To achieve the best test performance, the episodic pre-training strategy should closely mimic the test scenario. In other words, if support and query samples are to be drawn from different subjects during testing, the subject-independent sampling strategy should be employed during episodic pre-training. Meanwhile, if support and query samples are to be drawn from the same subject during testing, subject independent sampling strategy should be adopted over the other sampling strategies.
\par
Instead of matching time series, McCartney et al. \cite{mccartney2019zeroa} designed a multi-modal matching network for \ac{BCI} image retrieval. The task involves identifying the visual stimuli of a given \ac{EEG} signal. A variety of manually selected filters and correlation metrics were used to extract features from \ac{EEG} and corresponding visual stimuli images. Linear regression models are then trained to map \ac{EEG} features to visual stimulus features and determine the most likely visual stimulus through the nearest neighbor. Wang et al. \cite{wang2023generalized} propose to further extend the matching network to \ac{SSVEP} classification under a generalized zero-shot learning setting, where query samples can belong to both seen and unseen classes during training. \ac{SSVEP} describes the brain's natural response to visual stimulation of specific frequencies. For a given \ac{EEG} signal, the \ac{SSVEP} classification problem is to predict the modulation frequency of a visual stimulus, typically in the form of a flickering screen at set frequencies. The authors propose to replace the support set \ac{EEG} samples with the sine wave template of the luminance modulation function of the visual stimuli for all classes. Consequently, the matching network consists of two separate embedding networks, one for projecting query \ac{EEG} signals to feature vectors and the other for projecting a sine wave template of visual stimuli to feature vectors of the same representation space. Capitalizing on the use of a visual stimuli template and prior knowledge of the problem setup, the proposed matching network is capable of achieving multi-fold improvements to existing zero-shot or training-free methods, as well as comparable results to SOTA methods trained on the full dataset.

\subsubsection{Prototypical Network}
\begin{figure}
    \centering
    \includegraphics[width=0.45\textwidth]{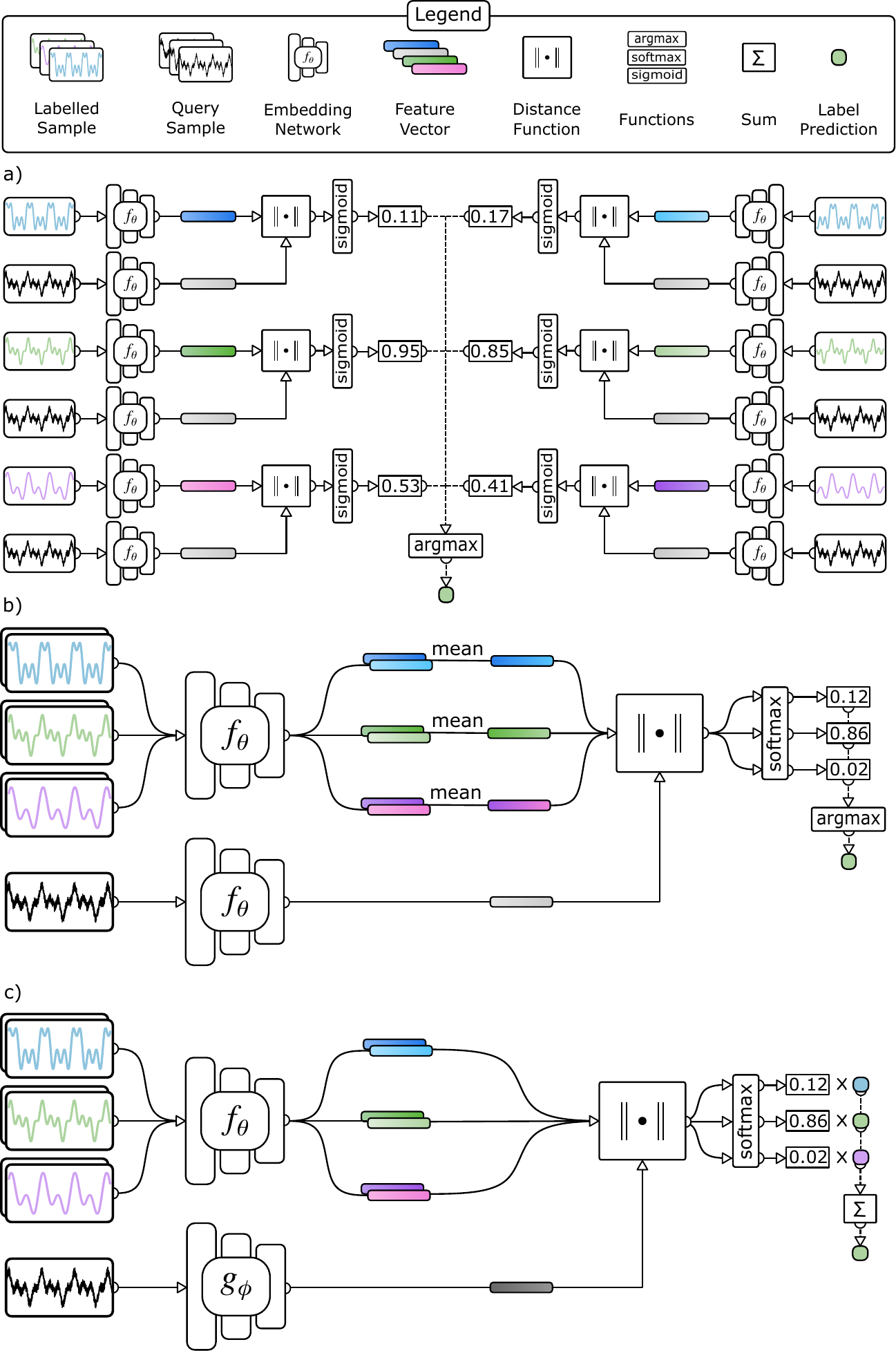}
    \caption{\textbf{Comparison of metric-based methods in few-shot learning with a 3-way-2-shot setup.} All methods employ an embedding network to project support and query samples into the representation space, in order to measure the similarity between support and query samples. a) Siamese network computes the similarity between each pair of support-query feature vectors to determine the most likely class. b) Prototypical network computes class-wise prototypes by taking the average of all support sample feature vectors from each class. Similarity is then computed between the class-wise prototypes and the query feature vector. c) Matching network directly computes the similarity between every pair of support and query feature vectors, but the support and query samples are projected using two different embedding networks. The network output is a linear combination of similarity between each support-query pair and class of the support sample.}
    \label{fig:metric}
\end{figure}

Prototypical networks \cite{snell2017prototypical} are similar to the matching networks, but instead of comparing the query embedding with every support embedding, the query embeddings are compared against class-wise prototype embeddings. Given a $N$-way-$K$-shot support set $\mathcal{S} = \{ (\mathbf{x}_{s,i}, y_{s,i}) \}_{i=1}^{N\times K}$, prototypical networks project support samples of a given class into embeddings of the feature space and construct class-wise prototypes by computing the average over the feature vectors. For class $k$, the prototype can be constructed as follows:
\begin{equation*}
    \mathbf{c}_k = \frac{1}{|\mathcal{S}_k|} \sum_{(\mathbf{x}_{s,i}, y_{s,i}) \in \mathcal{S}_k}{f_\theta(\mathbf{x}_{s,i})}
\end{equation*}
where $\mathcal{S}_k$ denotes the subset of $\mathcal{S}$ that belongs to class $k$, $|\mathcal{S}_k|$ denotes the number of samples of the subset, and $|\mathcal{S}_k| = N$ for $N$-way-$K$-shot support set $\mathcal{S}$. A similarity metric computes the similarity between the $j$-th query sample $\mathbf{x}_{q,j}$ against each of the $K$ prototypes $\mathbf{c}_k$, using a combination of distance function and softmax activation:
\begin{equation*} \label{eq:protopred}
    s(\mathbf{x}_{q,j}, \mathbf{c}_k) = \frac{\text{exp}(-||f_\theta(\mathbf{x}_{q,j}), \mathbf{c}_k||)}{\sum\limits_{k'=1}^{K}\text{exp}(-||f_\theta(\mathbf{x}_{q,j}), \mathbf{c}_{k'}||)}
\end{equation*}
The final prediction $\hat y_{q,j}$ for the $j$-th query sample $\mathbf{x}_{q,j}$ is then:
\begin{equation*}
    \hat y_{q,j} = \argmax_k s(\mathbf{x}_{q,j}, \mathbf{c}_k)
\end{equation*}
Prototypical networks are less computationally expensive than Siamese networks, as the similarity between the query feature vector is computed against $K$ prototypes in prototypical networks, as opposed to $N \times K$ supporting feature vectors in Siamese networks. In the case of one-shot learning, matching networks and prototypical networks are equivalent. The major difference between prototypical, matching, and Siamese networks lies in how the similarity between the support and query embeddings is computed. Siamese networks compute the distance between pairs of query and support embeddings. Matching networks compute the distance between query embedding and every support sample embedding. Prototypical networks compute the distance between query embedding and prototype embedding of each class. Figure \ref{fig:metric} provides a visual illustration of the differences between the metric-based methods using a 3-way-2-shot setup.
\par
The choice of embedding network and similarity metric for prototypical networks is similarly diverse. Salekin et al. \cite{salekin2021understanding} implemented a prototypical network based on 1D \ac{CNN} embedding network and Euclidean distance function. The authors conclude from the experimental results that, under a class-imbalanced few-shot learning setting,  prototypical networks do not suffer from overfitting like conventional supervised classifiers. Using a similar setup, Hernandez-Galvan et al. \cite{hernandez2023prototypical} adopted an embedding network that consists of a mixture of 1D convolution and \ac{GRU} layers, as well as softmax activated Euclidean distance function. Meanwhile, Tang et al. \cite{tang2020interpretable} chose to extract features using handcrafted symbolic Fourier transformation word histograms and a fully connected layer. A softmax-activated Euclidean distance function between the query embeddings and prototypes is chosen as the similarity metric. The authors further take advantage of the interpretability of the handcrafted features and prototypes to facilitate model explanation, data analysis, and problem understanding.  Despite the popularity of the Euclidean distance function as part of the similarity metric, Pan et al. \cite{pan2023multiplea} also adopted the cosine distance function and a multi-scale 1D \ac{CNN} as the embedding network to extract features of different scales and contextual characteristics. An ablation study found that cosine distance as a similarity metric led to better model performance than Euclidean distance, potentially due to the fact that cosine distance increases the distance between samples of different classes \cite{sun2022euler}. This is contrary to the findings when prototypical networks are first proposed, where the authors found that squared Euclidean distance greatly improves results and conjecture that cosine distance performs relatively poorly due to it not being a Bregman divergence \cite{snell2017prototypical}. Therefore, the choice of similarity metric plays an important role in the model's performance and could vary depending on many factors, such as the architecture of the embedding network.
\par
Hyperdimensional computing presents an alternative feature extractor to 1D \ac{CNN}-based embedding networks. Moin et al. \cite{moin2018emg} chose to extract features using a hyperdimensional computing framework that consists of three components: mapping module, encoder, and associative memory. The mapping module converts raw biosignals into hyperdimensional vectors, and the encoder module extracts spatial and temporal features using hyperdimensional vector space operations, such as point-wise multiplication, point-wise addition, scalar multiplication, and permutation. The associative memory module stores prototypes of encoded features for each class. During testing, the most similar hyperdimensional vector to the prototype hyperdimensional vectors is identified using cosine distance. Adopting a similar ideology, Burrello et al. \cite{burrello2019hyperdimensionala} combined hyperdimensional computing with \ac{LBP} to achieve near-perfect seizure detection performance with one and few-shot samples. Local \ac{LBP} representation of the brain state is constructed over time for all electrodes. Then, \ac{LBP} representations are manipulated by hyperdimensional operations to produce complex representation vectors, which form the basis for constructing prototypes of each class. Hamming distance is selected as the distance function for the similarity metric during inference to select the best matching prototype to the query sample. The use of \ac{LBP} brings transparency to the learning procedure. It allows for the translation of the learned codes into spatial localization of seizure-generating regions, which provides useful insights to support clinical decision-making. 
\par
Several studies explored alternative methods of constructing prototypes from feature vectors of the support set. Sun et al. \cite{sun2023few} argue that some relationship between prototypes exists, thus, inter-class relationships amongst the prototypes can improve model learning. Consequently, the author proposes a self-attention-based prototype enhancement module to generate enhanced prototypes that take into account the relationship among prototypes. Furthermore, to adapt to a class incremental few-shot learning setting, a prototype non-overlapping loss has been proposed to encourage the separation of prototypes of new classes from prototypes of old classes. Similarly, Suo et al. \cite{suo2020tadanet} explored learning the relationship amongst prototypes with supervision from a hierarchy graph. Simply taking the average of feature vectors from the embedding network as prototypes can be sensitive to outliers given limited support samples of each class. The authors observed that hierarchy graphs capture relationships amongst classes, where leaf node classes that share the same ancestral nodes are likely to share similar characteristics. Therefore, prototypes of leaf node classes that belong to the same ancestral node should be closer than prototypes of leaf node classes that belong to different ancestral nodes \cite{suo2020tadanet}. Thus, an attention-based procedure has been proposed to enhance prototype representations using information from the hierarchy graph. However, the availability of a hierarchy graph cannot be assured, limiting the applicability of this method to different applications. Instead of computing prototypes as the average across feature vectors of the support set, Ahn et al. \cite{ahn2023automatic} drew inspiration from \cite{wertheimer2021few} and constructed prototypes by pooling the feature vectors of each class from the support set as a feature matrix. The similarity metric between the feature matrix prototype and the feature vector of the query sample is calculated as a solution of ridge regression on the feature map.

\subsubsection{Summary}
Metric-based methods learn to compare, which results in flexible models that are not anchored to specific classes. By creating contrastive pairs and sets, metric-based methods are more robust to inter-subject variability and class imbalance than traditional supervised learning, as demonstrated through a variety of applications. While diverse combinations of embedding networks and similarity metrics have been proposed, there is a lack of consensus on how to pick the embedding networks and similarity metrics. 

\subsection{Model-Based Few-Shot Learning}
\begin{table*}
\renewcommand{\arraystretch}{1.05}
\centering
\caption{\textbf{Model-based few-shot learning methods for biomedical time series}}
\resizebox{1\textwidth}{!}{
\begin{tabular}{ccccccccccc} 
\toprule
\toprule
 & \multicolumn{3}{c}{\textbf{Problem}} & \multicolumn{1}{c}{\textbf{Pre-Training}} & \multicolumn{1}{c}{\begin{tabular}[c]{@{}c@{}}\textbf{Knowledge} \\ \textbf{Transfer}\end{tabular}} & \multicolumn{4}{c}{\textbf{Few-Shot Learning Setup}} & \multicolumn{1}{c}{\textbf{Open Access}} \\ \cmidrule(lr){2-4} \cmidrule(lr){5-5} \cmidrule(lr){6-6} \cmidrule(lr){7-10} \cmidrule(lr){11-11}
\textbf{Reference} & \multicolumn{1}{c}{Task} & \multicolumn{1}{c}{Modality} & \multicolumn{1}{c}{Method} & \multicolumn{1}{c}{\begin{tabular}[c]{@{}c@{}}Type of \\ Pre-Training\end{tabular}} & \multicolumn{1}{c}{\begin{tabular}[c]{@{}c@{}}Type of \\ Transfer\end{tabular}} & \multicolumn{1}{c}{FSL Dataset} & \multicolumn{1}{c}{\begin{tabular}[c]{@{}c@{}}Number of \\ Classes\end{tabular}} & \multicolumn{1}{c}{Support Set Size} & \multicolumn{1}{c}{Evaluation Metric} & \multicolumn{1}{c}{\begin{tabular}[c]{@{}c@{}}Code \\ Availability\end{tabular}} \\
\midrule
 \cite{phunruangsakao2022deep} & Motor Imagery & {EEG} & Adversarial & Supervised & Subject &\begin{tabular}[c]{@{}c@{}} BCI-IV 2a \cite{tangermann2012review}\\BCI-IV 2b \cite{tangermann2012review} \end{tabular}&\begin{tabular}[c]{@{}c@{}} 4\\2 \end{tabular}&\begin{tabular}[c]{@{}c@{}} ?\\? \end{tabular}& AC, CK & \ding{55} \\ \hline
 \cite{zhu2021unsupervised} & Seizure & {EEG} & Adversarial & Unsupervised & Subject & \cite{wagenaar2015collaborating} & 2 & \{0,1,2,3\}-shot & AUROC & \ding{55} \\ \hline
 \cite{zhang2022patienta} & Seizure & {EEG} & {SVM} & - & N/A &\begin{tabular}[c]{@{}c@{}} CHB-MIT \cite{shoeb2009application}\\Collected (Private) \end{tabular}&\begin{tabular}[c]{@{}c@{}} 2\\2 \end{tabular}&\begin{tabular}[c]{@{}c@{}} 1-shot\\1-shot \end{tabular}& SE, SP & \ding{55} \\ \hline
 \cite{ma2023fewa} & Emotion & {EEG} & Custom & Supervised & Classes & SEED-V \cite{liu2021comparing} & 5 & 5-shot & AC & \ding{55} \\ \hline
 \cite{pan2022few} & {HAR} & ACC, Video & Custom & Supervised, Episodic & Classes &\begin{tabular}[c]{@{}c@{}} \cite{song2016multimodal}\\Stanford ECM \cite{nakamura2017jointly} \end{tabular}&\begin{tabular}[c]{@{}c@{}} 5\\5 \end{tabular}&\begin{tabular}[c]{@{}c@{}} \{1,3\}-shot\\\{1,3\}-shot \end{tabular}& AC & \ding{55} \\ \hline
 \cite{hartmann2023high} & {HAR} & {EMG}, ACC & Custom & - & N/A &\begin{tabular}[c]{@{}c@{}} CSL-SHARE \cite{liu2021csl}\\UniMiB SHAR \cite{micucci2017unimib} \end{tabular}&\begin{tabular}[c]{@{}c@{}} 22\\17 \end{tabular}&\begin{tabular}[c]{@{}c@{}} \{1-500\}-shot\\\{1-500\}-shot \end{tabular}& AC & \ding{55} \\ \hline
 \cite{wang2023personalized} & Blood Pressure & {PPG} & {SVM} & - & N/A & Private (Collected) & Regression & \{10,15,20,30,40,45\}-shot & MAE, MPE, PC, SD & \ding{55} \\ \hline
 \cite{duan2020zero} & Motor Imagery & {EEG} & Custom & Supervised & Classes & Private (collected) & 3 (1 unseen) & 0-shot & AC & \ding{55} \\ \hline
 \cite{rahimian2021fewa} & Hand Gesture & {EMG} & Custom & Supervised, Episodic &\begin{tabular}[c]{@{}c@{}} Repetitions\\Subjects\\Classes \end{tabular}& Ninapro DB2 \cite{pizzolato2017comparison} &\begin{tabular}[c]{@{}c@{}} 50 (5-way)\\50 (5-way)\\34 (5-way) \end{tabular}&\begin{tabular}[c]{@{}c@{}} \{1,5\}-shot\\\{1,5\}-shot\\\{1,5\}-shot \end{tabular}& AC & \ding{55} \\ \hline
\bottomrule
\end{tabular}%
}
\raggedright
\begin{tablenotes}
All tasks, except blood pressure prediction, are classification tasks. Abbreviations: Electroencephalogram (EEG), Accelerometry (ACC), Electromyography (EMG), Photoplethysmography (PPG), Accuracy (AC), Cohen's Kappa (CK), Area Under Receiver Operating Characteristic Curve (AUROC), Sensitivity (SE), Specificity (SP), Mean Absolute Error (MAE), Mean Percentage Error (MPE), Pearson Correlation (PC), Standard Deviation of Estimation Error (SD)
\end{tablenotes}
\end{table*}
Model-based few-shot learning methods rely on designing a model architecture that can generalize well to different few-shot learning tasks.

\subsubsection{Adversarial Networks}
Adversarial learning-based models design modules that compete against each other and improve together during the learning process. While \ac{GAN} presented in Section \ref{sec:data-gan} is a subset of adversarial learning-based models, research works presented in this section focus on designing adversarial models that generalize across domains, as opposed to supplementing the few-shot support set with synthetic data or features.
\par
Phunruangsakao et al. \cite{phunruangsakao2022deep} propose an adversarial domain adaptation framework that guides the model to extract subject-invariant features. The proposed framework first pre-trains a classification model using samples from the source subjects. The pre-trained model is then frozen and queried using samples from both the source and target subjects. The features extracted by the \ac{CNN} backbone of the classification model are provided as input to a discriminator, which attempts to distinguish the origin of the feature as the source or target subject. To train the discriminator and the backbone network of the classification model in an adversarial manner, the weights of one network are frozen while the weights of the other are updated. This process encourages the \ac{CNN} backbone to extract features that are invariant between the source and the target subjects, effectively enabling positive transfer that adapts knowledge from the source domain to boost performance in the few-shot target domain.
Zhu et al. \cite{zhu2021unsupervised} adopt a similar adversarial domain adaptation framework in an unsupervised manner. Instead of pre-training on source data and asynchronously updating the discriminator and backbone network, Zhu et al. propose to train a discriminator and an autoencoder simultaneously. The discriminator learns to distinguish between source and target subject samples using the latent space representation from the encoder. At the same time, a reconstruction loss guides the training of the autoencoder. Thus, the encoder is encouraged to extract subject-invariant features and enables the direct application of a classifier trained on source subject data to the target subject.

\subsubsection{\ac{SVM}}
\ac{SVM} is a class of supervised machine learning algorithms that attempts to find the optimal decision boundary or hyperplane that separates different classes within a given feature space. The optimal hyperplane maximizes the distance between the hyperplane and the closest data points to the hyperplane of each class, commonly referred to as support vectors. While \ac{SVM} hyperplane can only handle linear decision boundaries, the use of "kernel trick" to project input features to high dimensional space enables the separation of non-linear patterns. The concept of \ac{SVM} can be further expanded to regression problems, also known as \ac{SVR}. Instead of treating the hyperplane as a decision boundary that maximizes the distance to data points of each class, the hyperplane serves to fit the data points. The "kernel trick" can be applied similarly to fit non-linear regression problems.
\par
In \cite{zhang2022patienta}, Zhang et al. propose an epilepsy \ac{SoC} consisting of a time-channel averaging feature extractor and a \ac{SVM} classifier with a second-order polynomial guided kernel. One-shot learning in a patient recruited from a local hospital achieved a vector-based sensitivity of 39.5\% and a specificity of 98. 9\% on seven seizures in four hours. Online tuning using three additional false negatives and one false positive as support vectors further boosted the vector-based sensitivity to 71.9\%. 
In another application, a modified \ac{SVR} algorithm was proposed for blood pressure regression using \ac{PPG} signals \cite{wang2023personalized}. Namely, an error feedback model fine-tuning mechanism is incorporated into the \ac{SVR} algorithm, which achieves good few-shot learning performance and can improve the long-term monitoring capability by adapting to new samples without model retraining. Using the proposed framework, a personalized blood pressure prediction model can be constructed with few samples and continue to model blood pressure variations after three months of constructing the model.
\ac{SVM} remains a popular choice for few-shot learning in wearable and edge devices due to its low memory \& computation burden as well as its ease of online tuning with additional samples. While deep neural network-based few-shot learning methods achieve SOTA performance for cross-dataset, cross-task performance, \ac{SVM} excels at building patient-specific models, especially when training from scratch is necessary due to a lack of access to external data sources to facilitate knowledge transfer. A major limitation of \ac{SVM} is its reliance on manual feature selection, which is laborious and also heavily influences the performance of the model.
\subsubsection{Customized Models and Frameworks}
In addition to existing models, many researchers have also explored novel, customized frameworks to achieve few-shot learning. Some focused on designing feature extraction and classifier modules that can generalize with few-shot samples and others re-imagined the classification framework in its entirety.
For example, Pan et al. \cite{pan2022few} explored graph networks for multi-modal egocentric activity recognition. To extract effective features and recognize new classes from few-shot multi-modal samples, a two-stream graph network was proposed. The heterogeneous graph-based multi-modal association module captures dynamic and complementary information from multi-modal data. The knowledge-aware classifier module captures semantic relations between activity classes and objects to assist generalization to unseen classes. An ablation study of different modules shows that the heterogeneous graph network is useful for fusing multi-modal information and the knowledge-aware classifier shows improvements over the linear classifier. Comparison against existing baselines with the likes of \ac{LSTM}, matching networks, and prototypical networks further demonstrate the effectiveness of the approach. 
\par
Incremental few-shot learning was explored by Ma et al. \cite{ma2023fewa} for emotion recognition. The key idea behind the framework is to add new weights to the final linear classifier for each new class introduced by the incremental learning stages. During the base stage, the \ac{GCN} feature extractor and a linear classifier are pre-trained using abundant data from the base classes. During incremental stages, the \ac{GCN} feature extractor and linear classifier weights from the previous stages are frozen. New linear classifier weights for incremental classes are added and fine-tuned with the support set. Entropy and subspace regularization were introduced to help the model learn from new samples and avoid catastrophic forgetting of old classes during the few-shot fine-tuning process.
On a different note, Duan et al. \cite{duan2020zero} explored zero-shot outlier detection based on local density. The proposed framework consists of two stages: projection to target space and novelty detection. In the first stage, a common spatial pattern feature extractor and a two-layer fully connected network are used to project EEG time series to semantic feature space. In the second stage, outlier classification is done in a zero-shot manner by determining whether the features belong to an unseen class, based on the feature manifold of seen classes. Without any information on the unseen class, the proposed method can achieve accuracy similar to that of a classifier trained with full access to all classes.
\par
Delving into an alternative perspective, Hartmann et al. \cite{hartmann2023high} propose a few-shot learning framework based on high-level features. Individual classes are defined through a combination of different high-level features, and separate classifiers are trained to provide binary classification for each of the high-level features. With the pre-trained classifiers for high-level features, zero-shot learning for new classes becomes a natural extension of the framework. An inherent advantage of the framework lies within its model interpretability and ease of performing error attribution analysis. However, this comes at the cost of time and domain knowledge required to define such high-level features. The extension of zero-shot learning beyond the finite combination of high-level features presents another shortcoming of the framework.
In another investigation, Rahimian et al. \cite{rahimian2021fewa} propose an architecture for processing support and query samples similar to language models, where the input of the model receives concatenated support samples with labels followed by query samples with null labels. The network consists of attention and dilated causal 1D convolution modules to process the relationship between inputs. Extensive experiments show that the network can generalize to new classes and new repetitions, although there is a lack of comparison against existing baselines and few-shot methods. 
\subsubsection{Summary}
Model-based methods approach the design of model architecture from creative perspectives. Adversarial networks excel at learning domain- and subject-invariant features, while \acp{SVM} are capable of learning from few-shot samples without any pre-training. Customized frameworks provide additional functionalities such as incremental learning, multi-modal learning, and outlier detection, which more closely mimics how humans learn. However, many model-based frameworks make strong assumptions about problem setup, which does not hold when applied to other applications.

\subsection{Optimization-Based Few-Shot Learning}
\begin{table*}
\renewcommand{\arraystretch}{1.04}
\centering
\caption{\textbf{Optimization-based few-shot learning methods for biomedical time series}}
\resizebox{1\textwidth}{!}{
\begin{tabular}{ccccccccccc} 
\toprule
\toprule
 & \multicolumn{3}{c}{\textbf{Problem}} & \multicolumn{1}{c}{\textbf{Pre-Training}} & \multicolumn{1}{c}{\begin{tabular}[c]{@{}c@{}}\textbf{Knowledge} \\ \textbf{Transfer}\end{tabular}} & \multicolumn{4}{c}{\textbf{Few-Shot Learning Setup}} & \multicolumn{1}{c}{\textbf{Open Access}} \\ \cmidrule(lr){2-4} \cmidrule(lr){5-5} \cmidrule(lr){6-6} \cmidrule(lr){7-10} \cmidrule(lr){11-11}
\textbf{Reference} & \multicolumn{1}{c}{Task} & \multicolumn{1}{c}{Modality} & \multicolumn{1}{c}{Method} & \multicolumn{1}{c}{\begin{tabular}[c]{@{}c@{}}Type of \\ Pre-Training\end{tabular}} & \multicolumn{1}{c}{\begin{tabular}[c]{@{}c@{}}Type of \\ Transfer\end{tabular}} & \multicolumn{1}{c}{FSL Dataset} & \multicolumn{1}{c}{\begin{tabular}[c]{@{}c@{}}Number of \\ Classes\end{tabular}} & \multicolumn{1}{c}{Support Set Size} & \multicolumn{1}{c}{Evaluation Metric} & \multicolumn{1}{c}{\begin{tabular}[c]{@{}c@{}}Code \\ Availability\end{tabular}} \\
\midrule
 \cite{wu2022does} & Motor Imagery & {EEG} & {MAML} & Supervised, Episodic & Subjects &\begin{tabular}[c]{@{}c@{}} BCI-IV 2a \cite{tangermann2012review}\\\cite{cho2017eeg}\\PhysionetMI \cite{schalk2004bci2000} \end{tabular}&\begin{tabular}[c]{@{}c@{}} 4\\2\\4 \end{tabular}&\begin{tabular}[c]{@{}c@{}} \{1,5\}-shot\\\{1,5\}-shot\\\{1,5\}-shot \end{tabular}& AC & \ding{55} \\ \hline
 \cite{lan2020gazegraph} & Gait Impairment & {VOG} & {MAML} & Supervised, Episodic & Subjects &\begin{tabular}[c]{@{}c@{}} SedentaryActivity \cite{srivastava2018combining}\\JapaneseDocument \cite{kunze2013know}\\Collected \cite{lan2020gazegraph} \end{tabular}&\begin{tabular}[c]{@{}c@{}} 6\\8\\5 \end{tabular}&\begin{tabular}[c]{@{}c@{}} \{5,10\}-shot\\\{5,10\}-shot\\\{5,10\}-shot \end{tabular}& F1, AC & \ding{55} \\ \hline
 \cite{li2022meta} &\begin{tabular}[c]{@{}c@{}} Motor Imagery\\Motor Imagery\\{ERP}\\{ERP} \end{tabular}& {EEG} & {MAML} & Supervised, Episodic & Subjects &\begin{tabular}[c]{@{}c@{}} BCI-IV 2a \cite{tangermann2012review}\\BNCI-Horizon 002-2014 \cite{tangermann2012review}\\BNCI-Horizon 009-2014 \cite{arico2014influence}\\BCI Challenge \cite{mattout2015bci} \end{tabular}&\begin{tabular}[c]{@{}c@{}} 4\\2\\2\\2 \end{tabular}&\begin{tabular}[c]{@{}c@{}} \{0,2,4,6\}-shot\\\{0,2,4,6\}-shot\\\{0,2,4,6\}-shot\\\{0,2,4,6\}-shot \end{tabular}& AUROC, AC & \href{https://github.com/sylyoung/MetaEEG}{\ding{51}} \\ \hline
 \cite{zhang2019metapred} & Clinical Risk & {EHR} & {MAML} & Supervised, Episodic & Tasks & OHSU Hospital (Private) & 2 & 20\% & AUROC, F1 & \href{https://github.com/sheryl-ai/MetaPred}{\ding{51}} \\ \hline
 \cite{youssef2022model} & Blood Pressure & {EHR} & {MAML} & Supervised, Episodic & Subjects & MIMIC-III \cite{johnson2016mimic} & Regression & 10-shot & RMSE, MAE & \href{https://github.com/paulyoussef/model-personalization}{\ding{51}} \\ \hline
 \cite{meyer2022u} & {HAR} & {IMU}, {VOG} & TL & Supervised, Episodic & Subjects & Collected (Private) & 7 & 3-shot & AC, Macro-F1 & \ding{55} \\ \hline
 \cite{lv2022exploratory} & Gait & {EEG}, {EMG} & TL & Supervised & Subject & Collected (Private) & 4 & 20\% & AC & \ding{55} \\ \hline
 \cite{nazari2022epilepsya} & Seizure & {EEG} & TL & Supervised & Subjects & CHB-MIT \cite{shoeb2000chb} & 2 & 3-shot & SE, FPR, AUROC & \ding{55} \\ \hline
 \cite{hur2023genhpf} & ICU Outcome & {EHR} & TL & Semi-Supervised & Dataset &\begin{tabular}[c]{@{}c@{}} MIMIC-III \cite{johnson2016mimic}\\eICU \cite{pollard2018eicu}\\MIMIC-IV \cite{johnson2020mimic} \end{tabular}&\begin{tabular}[c]{@{}c@{}} Varying\\Varying\\Varying \end{tabular}&\begin{tabular}[c]{@{}c@{}} \{0,10,30,50,70\}\%\\\{0,10,30,50,70\}\%\\\{0,10,30,50,70\}\% \end{tabular}& AUROC & \href{https://github.com/hoon9405/GenHPF}{\ding{51}} \\ \hline
 \cite{bhaskarpandit2023how} & Myocardial Infarction & {ECG} & TL & Supervised & Tasks & \cite{khan2021ecg} & 3 &\begin{tabular}[c]{@{}c@{}} \{0,10,30,50,70\}\%\\\{0,10,30,50,70\}\%\\\{0,10,30,50,70\}\% \end{tabular}& AC & \ding{55} \\ \hline
 \cite{saeed2021sense} &\begin{tabular}[c]{@{}c@{}} {HAR}\\{HAR}\\{HAR}\\{HAR}\\{HAR}\\Sleep Staging\\Stress \end{tabular}&\begin{tabular}[c]{@{}c@{}} ACC, GYR\\ACC, GYR\\ACC, GYR\\ACC, GYR\\ACC, GYR\\EEG, EOG\\HR, SC \end{tabular}& TL & Self-Supervised & Datasets &\begin{tabular}[c]{@{}c@{}} HHAR\\MobiAct\\MotionSense\\UCI HAR\\HAPT\\Sleep-EDF\\MIT DriverDb \end{tabular}&\begin{tabular}[c]{@{}c@{}} 6\\11\\6\\6\\12\\5\\2 \end{tabular}&\begin{tabular}[c]{@{}c@{}} \{5,10\}-shot\\\{5,10\}-shot\\\{5,10\}-shot\\\{5,10\}-shot\\\{5,10\}-shot\\\{5,10\}-shot\\\{5,10\}-shot \end{tabular}& F1 & \ding{55} \\ \hline
\bottomrule
\end{tabular}%
}
\raggedright
\begin{tablenotes}
All tasks, except blood pressure prediction, are classification tasks. Abbreviations: Electroencephalogram (EEG), Video-Oculograph (VOG), Electronic Health Record (EHR), Inertial Measurement Unit (IMU), Accelerometry (ACC), Gyroscope (GYR), Electrooculography (EOG), Heart Rate (HR), Skin Conductance (SK), Accuracy (AC), Area Under Receiver Operating Characteristic Curve (AUROC), Root Mean Squared Error (RMSE), Mean Absolute Error (MAE), Sensitivity (SE), False Positive Rate (FPR). \\
\end{tablenotes}
\end{table*}
Optimization-based few-shot learning methods focus on developing innovative ways to train models. The non-convex nature of deep neural networks means that there are many local optimums that the model weights can converge to. Optimization-based methods focus on designing the optimization objective and training procedure to guide the model to converge to weight spaces that generalize well to new tasks when fine-tuned with few-shot samples. This family of methods is closely associated with meta-learning, learning how to learn.
\subsubsection{Transfer Learning}
Transfer learning aims to reuse a model trained on one task and adapt it to a related task. Typically, a model is pre-trained on the source task and subsequently fine-tuned on the target task with a relatively smaller dataset. Model-level transfer refers to fine-tuning the entirety of the model, and it assumes that the input and output structures of the source and target tasks are identical. Feature-level transfer refers to freezing feature extractors and fine-tuning task-specific heads. Pre-training on large source datasets helps the model to learn meaningful and generalizable features across tasks, providing an initialization to boost performance on the target task.
\par
Feature-level transfer learning is a popular method amongst few-shot learning applications to biomedical time series. Meyer et al. \cite{meyer2022u} applied transfer learning to a 1D UNet for model personalization. To adapt to new subjects, the pre-trained encoder and decoder are frozen while the classifier with fully connected layers is re-initialized and trained using a few samples from the new subject. Training using only datasets from the source subject, the proposed 1D UNet outperforms the baseline \ac{SVM} on the targeted subject test set. Its performance is further boosted when fine-tuned with few-shot samples from the target subject in question. Nazari et al. \cite{nazari2022epilepsya} studied the choice of classifier in the performance of few-shot transfer learning for seizure detection. After pre-training the 2D \ac{CNN} backbone and classifier using data from 15 source subjects, fine-tuning the \ac{SVM} classifier led to a superior performance on the target subject than using a fully connected layer as a classifier. Bhaskarpandit et al. \cite{bhaskarpandit2023how} provided a benchmark on the number of samples needed for transfer learning to be effective when applied to myocardial infarction classification. A variety of pre-trained backbone networks were studied, including VGG16 \cite{simonyan2014very}, DenseNet \cite{huang2017densely}, InceptionV3 \cite{szegedy2016rethinking}, ResNet \cite{he2015delving}, and EfficientNet \cite{tan2019efficientnet}. Unsurprisingly, as the number of support samples for fine-tuning increased, the few-shot learning performance improved. With around 5 or fewer support samples, transfer learning fails, but with around 20 support samples, performance surpasses 99\% accuracy and the marginal benefit of additional samples diminishes. 
\par
Other works have attempted to modify the pre-training process to facilitate fine-tuning. Lv et al. \cite{lv2022exploratory} studied two types of pre-training strategies for transfer learning: cross-subject mixture transfer and model averaging transfer. Cross-subject mixture transfer mixes all source subjects' data for pre-training, while model averaging transfer averages the weights of models individually trained on each source subject's data. Experiments were carried out with a 2D \ac{CNN} consisting of two convolution layers and a fully connected layer. The results show that cross-subject mixture transfer is useful and provides the best performance boost when the fully connected layers are fine-tuned, which agrees with the established norm of fine-tuning only the classifier and not the feature extractors. On the other hand, model averaging transfer provides worse performance than cross-subject mixture transfer, despite its similarity to the FedAvg \cite{mcmahan2017communication} algorithm, widely adopted as a baseline for federated learning. Hur et al. \cite{hur2023genhpf} propose to extend transfer learning to a multi-task, multi-source setting and pre-train the source model in a self-supervised manner. To overcome the heterogeneity of medical codes and schemes of \acp{EHR} used by different hospitals and datasets, the authors propose to convert \acp{EHR} to text representation, thus eliminating the pre-processing and feature engineering procedure for each hospital. Then, multi-source pre-training with unlabeled data is performed using different self-supervised learning strategies: SimCLR \cite{chen2020simple}, Wav2Vec 2.0 \cite{baevski2020wav2vec}, MLM \cite{devlin2018bert} and SpanMLM \cite{joshi2020spanbert}. Few-shot samples from the target hospital or dataset enable fine-tuning of the multi-source model, adapting to any hospital without restraints on the data format. The authors found that transfer learning with self-supervised pre-training methods outperforms training from scratch, except for SpanMLM. Furthermore, SimCLR is the best self-supervised pre-training method, which focuses on learning patient-level representations, as opposed to event-level or token-level representations with other self-supervised pre-training strategies. In addition, multi-source pre-training was found to yield better transfer learning performance than single-source pre-training. Similarly, Saeed et al. \cite{saeed2021sense} also applied self-supervised pre-training to facilitate transfer learning. The authors introduce a variety of self-supervised pre-training tasks to encode underlying signal semantics of multi-modal time series from different sensor modalities. Examples of self-supervised pre-training tasks include differentiating between clean and blended signals, feature prediction from masked windows, and signal transformation recognition. Experimental results show that self-supervised networks serve as great initialization to boost performance with limited labeled data. For some datasets, self-supervised pre-training can outperform a fully supervised model trained with the entire dataset. Moreover, learned representations from self-supervised training are highly transferable across related datasets.

\subsubsection{\ac{MAML}}
\ac{MAML} \cite{finn2017model} was first proposed by Finn et al. in 2017 as a framework to initialize model parameters that are easy to fine-tune. The problem consists of a set of $T$ tasks $\mathcal{T} = \{ \mathcal{T}_t  \}_{t=1}^T$ and corresponding set of datasets $\mathcal{D} = \{ \mathcal{D}_t\}_{t=1}^T$ with each $\mathcal{D}_t = \{(\mathbf{x}_l, y_l)\}_{l=1}^{L_t}$ consisting of $L_t$ input-label pairs. \ac{MAML} aims to find an initialization for model parameters $\theta$ that can be adapted quickly to perform well on any arbitrarily chosen task from $\mathcal{T}$, after fine-tuning with few-shot samples from its corresponding dataset from $\mathcal{D}$. The pseudo-code for the \ac{MAML} algorithm is outlined in Algorithm \ref{alg:MAML}. 

\begin{algorithm}
\caption{Algorithm for \ac{MAML} \cite{finn2017model}}\label{alg:MAML}
\begin{algorithmic}[1]
\Require $\mathcal{T} = \{ \mathcal{T}_1, \mathcal{T}_2, ..., \mathcal{T}_T  \}$ set of tasks
\Require $\mathcal{D} = \{ \mathcal{D}_1, \mathcal{D}_2, ..., \mathcal{D}_T \}$ set of corresponding datasets
\Require $p(\mathcal{T})$ distribution over tasks
\Require $\alpha, \beta$ step size hyperparameters
\Require $I, J, B$ batch size hyperparameters
\State randomly initialize $\theta$
\While{not done}
\State Sample $B$ tasks from $\mathcal{T}$ following $p(\mathcal{T})$
\For{$t \gets 1$ to $B$}
    \State Sample $\mathcal{D}^{inner}_t = \{(\mathbf{x}_{i}, y_{i})\}_{i=1}^I $ from $\mathcal{D}_t$
    \State Evaluate $\mathcal{L}_{\mathcal{T}_t}$ and $\nabla_{\theta} \mathcal{L}_{\mathcal{T}_t}(f_{\theta})$ using $\mathcal{D}^{inner}_t$
    \parState{%
    Compute adapted parameters with gradient descent: $\theta'_t = \theta - \alpha \nabla_\theta \mathcal{L}_{\mathcal{T}_t}(f_{\theta})$}
    \parState{%
    Sample $\mathcal{D}^{outer}_t = \{(\mathbf{x}_{j}, y_{j})\}_{j=1}^J$ from $\mathcal{D}_t$}
\EndFor
\parState{%
Update $\theta \leftarrow \theta - \beta \nabla_\theta \sum\limits_{t=1}^{B}{\mathcal{L}_{\mathcal{T}_t}(f_{\theta'_t})}$ using $\mathcal{D}^{outer}_t$}
\EndWhile
\end{algorithmic}
\end{algorithm}

In general, \ac{MAML} consists of two weight update computations. \ac{MAML} starts by sampling a batch of $B$ tasks from the distribution over all tasks $p(\mathcal{T})$. For the $t$-th sampled task $\mathcal{T}_t$, the inner loop computes adapted parameters using a small number of support samples $\mathcal{D}^{inner}_t = \{(\mathbf{x}_{i}, y_{i})\}_{i=1}^I $ drawn from $\mathcal{D}_t$ where $(\mathbf{x}_{i}, y_{i})$ are the $i$-th input label pair of $\mathcal{D}^{inner}_t$. The adapted parameter for each task $\mathcal{T}_t$ can be computed as:
\begin{equation*}
    \theta_t' = \theta - \alpha \nabla_{\theta} \mathcal{L}_{\mathcal{T}_t}(f_\theta)
\end{equation*}
Although the formulation in the algorithm indicates that the adapted parameters involve a single gradient update step, the adapted model parameters can be obtained after taking several gradient steps in the inner loop. The outer loop then updates the model parameters using another set of small samples $\mathcal{D}^{outer}_t = \{(\mathbf{x}_{j}, y_{j})\}_{j=1}^J $ drawn from $\mathcal{D}_t$ where $(\mathbf{x}_{j}, y_{j})$ are the $j$-th input label pair of $\mathcal{D}^{outer}_t$, to enhance the model's generalization ability across all tasks.  This is also known as the meta-update step, which updates the model using the gradient over the loss of the adapted parameters across all sampled tasks. The meta-update can be written as: 
\begin{equation*}
    \theta \leftarrow \theta - \beta \nabla_\theta \sum_{t=1}^{B} {\mathcal{L}_{\mathcal{T}_t}(f_{\theta'_t})}
\end{equation*}
Note that the loss of the inner loop weight update is computed using the meta-model $\theta$ on each of the sampled datasets $\mathcal{D}^{inner}_t$ for each task $\mathcal{T}_t$, while the outer loop weight update is computed using the adapted model parameters $\theta_t'$ on another sampled dataset $\mathcal{D}^{outer}_t$ over all sampled tasks. Since the inner loop computation of $\theta'_t$ involves gradients over $\theta$, the outer loop computation requires second-order gradients. To reduce the computational complexity, \ac{FOMAML} that omits the second-order gradient computations was found to achieve nearly the same performance as \ac{MAML}. Drawing inspiration from \ac{FOMAML}, Reptile \cite{nichol2018first} proposes to use the difference in parameter weights between the adapted parameters and meta-parameters as the gradient for meta-update. Theoretical analysis and experimental results show that first-order meta-learning algorithms can achieve similar performance as \ac{MAML} with full second-order gradients.

\ac{MAML} and Reptile have been directly applied to several biomedical time series applications. Early efforts by Lan et al. \cite{lan2020gazegraph} proposed a spatial-temporal gaze graph-based model for \ac{VOG} and trained it using the \ac{MAML} paradigm for quick adaptation to few-shot samples from new subjects. In \ac{MAML} terminology, the authors treat each task as a randomly sampled subset of the source dataset, consisting of $N$-way-$K$-shot support and query sets that contain disjoint samples. This trains the model in a way that is easy to fine-tune and adapt to new datasets from new subjects, such that when it is fine-tuned on an unseen support set, it can achieve good performance on the unseen query set. Experimental results show that training the proposed gaze graph model using \ac{MAML} enables superior adaptation performance to the few-shot support and query set than a model trained using transfer learning. Similarly, Youssef et al. \cite{youssef2022model} benchmarked \ac{MAML} against transfer learning for blood pressure prediction. Unlike \cite{lan2020gazegraph}, each task consists of data from a specific subject. In particular, the authors found that given a fixed model architecture and pre-training dataset, pre-training using \ac{MAML} and pre-training using a traditional approach achieves the same performance on an unseen target subject without fine-tuning. However, after fine-tuning both models with few samples from the target subject, \ac{MAML} pre-trained model significantly outperforms the traditional pre-trained model. This reiterates the fact that \ac{MAML} does not necessarily pre-train a model that achieves good performance on all tasks it was pre-trained on, but rather a model that is easy to fine-tune and adapt quickly to new tasks. This is particularly important for applications like model personalization to address the challenges associated with inter-subject variability. In contrast, Wu et al. \cite{wu2022does} applied Reptile to pre-train EEGNet \cite{lawhern2018eegnet} and found no improvement over pre-training EEGNet using traditional one-step model parameter update. The authors hypothesize that it is possibly due to the fact that biomedical time series signals such as \ac{EEG} have fewer samples and higher variance than computer vision and natural language processing datasets, which makes it more challenging to learn representations that are transferable between subjects. A mere application of meta-learning algorithms would not suffice for all biomedical time series applications.
\par
Several works have also proposed to modify the \ac{MAML} framework for better adaptation to biomedical applications. Li et al. \cite{li2022meta} propose a modified version of \ac{MAML}, where the inner loop takes a few gradient steps based on data from one subject and calculates the meta-loss based on another. The inner loop is repeated for randomly paired subjects from the entire dataset before a meta-step is taken in the outer loop using the average of meta-loss from the inner loop. The modification of \ac{MAML} to take into account differences across subjects enables better few-shot adaptation to new subjects compared to existing baselines like prototypical networks \cite{snell2017prototypical} and the original \ac{MAML} \cite{finn2017model}. Furthermore, models trained using the proposed \ac{MAML} achieve better model calibration than models trained using the original \ac{MAML}, where the model's predicted probabilities better reflect the true likelihood of the model's prediction. Zhang et al. \cite{zhang2019metapred} propose another variation of \ac{MAML}. In the computation of gradients for meta-update, the authors propose to incorporate the inner loop loss for the source tasks, in addition to the loss on target task observed in traditional \ac{MAML}. The experimental results show that the proposed modification of \ac{MAML} outperforms the original \ac{MAML}, and the authors conclude that the incorporation of supervised knowledge from the source domain enhances meta-learning knowledge transfer.
\par
\subsubsection{Summary}
Optimization-based methods are highly flexible in terms of the model architecture, which allows them to be wrapped around any existing model. On the other hand, optimization-based methods rely heavily on access to extensive external pre-training datasets. The pre-training process can be computationally heavy and hyper-parameter sensitive, especially for meta-learning methods that involve inner- and outer-loop optimizations.

\subsection{Hybrid Methods}
\begin{table*}
\renewcommand{\arraystretch}{1.05}
\centering
\caption{\textbf{Hybrid few-shot learning methods for biomedical time series}}
\resizebox{1\textwidth}{!}{
\begin{tabular}{ccccccccccc} 
\toprule
\toprule
 & \multicolumn{3}{c}{\textbf{Problem}} & \multicolumn{1}{c}{\textbf{Pre-Training}} & \multicolumn{1}{c}{\begin{tabular}[c]{@{}c@{}}\textbf{Knowledge} \\ \textbf{Transfer}\end{tabular}} & \multicolumn{4}{c}{\textbf{Few-Shot Learning Setup}} & \multicolumn{1}{c}{\textbf{Open Access}} \\ \cmidrule(lr){2-4} \cmidrule(lr){5-5} \cmidrule(lr){6-6} \cmidrule(lr){7-10} \cmidrule(lr){11-11}
\textbf{Reference} & \multicolumn{1}{c}{Task} & \multicolumn{1}{c}{Modality} & \multicolumn{1}{c}{Method} & \multicolumn{1}{c}{\begin{tabular}[c]{@{}c@{}}Type of \\ Pre-Training\end{tabular}} & \multicolumn{1}{c}{\begin{tabular}[c]{@{}c@{}}Type of \\ Transfer\end{tabular}} & \multicolumn{1}{c}{FSL Dataset} & \multicolumn{1}{c}{\begin{tabular}[c]{@{}c@{}}Number of \\ Classes\end{tabular}} & \multicolumn{1}{c}{Support Set Size} & \multicolumn{1}{c}{Evaluation Metric} & \multicolumn{1}{c}{\begin{tabular}[c]{@{}c@{}}Code \\ Availability\end{tabular}} \\
\midrule
\cite{akbari2021meta} & Modality Translation & {ECG}, {Bio-Z} & Model + Optimization & Supervised, Episodic & Subjects & Private & N/A & \{2,5\}-shot & PRD, WDD, PC & \ding{55} \\ \hline
 \cite{li2023novel} &\begin{tabular}[c]{@{}c@{}} {ERP}\\Emotion\\Sleep Staging \end{tabular}& {EEG} & Metric + Optimization & Supervised, Episodic & Subjects &\begin{tabular}[c]{@{}c@{}} ERP \cite{hoffmann2008efficient}\\SEED \cite{duan2013differential}\\SleepEDF \cite{kemp2000analysis} \end{tabular}& N/A &\begin{tabular}[c]{@{}c@{}} 5-shot\\10-shot\\10-shot \end{tabular}& AC & \ding{55} \\ \hline
 \cite{aldahr2022addressinga} & Seizure & {EEG} & Data + Metric & Supervised & Subjets &\begin{tabular}[c]{@{}c@{}} Bonn \cite{andrzejak2001indications}\\CHBMIT \cite{shoeb2000chb} \end{tabular}&\begin{tabular}[c]{@{}c@{}} 3\\3 \end{tabular}& 60\% & RE, SP, AC, AUROC & \ding{55} \\ \hline
 \cite{kim2022automatic} & Arrhythmia & {ECG} & Data + Model & Supervised & Dataset &\begin{tabular}[c]{@{}c@{}} MIT-BIH \cite{goldberger2000physiobank}\\CinC DB \cite{clifford2017af} \end{tabular}&\begin{tabular}[c]{@{}c@{}} 4\\2 \end{tabular}&\begin{tabular}[c]{@{}c@{}} \{0,5,10,20\}-shot\\\{0,5,10,20\}-shot \end{tabular}& PR, RE, SP, F1, GM & \ding{55} \\ \hline
 \cite{moon2022explainable} & Gait & Pressure & Metric + Model & Semi-Supervised & Subjects & Collected \cite{moon2022explainable} & 2 & 10-shot & TNR, TPR, AC & \ding{55} \\ \hline
 \cite{liu2021few} & Arrhythmia & {ECG} & Metric + Optimization & Supervised & Dataset &\begin{tabular}[c]{@{}c@{}} MIT-BIH \cite{goldberger2000physiobank}\\PTB \cite{bousseljot1995nutzung} \end{tabular}&\begin{tabular}[c]{@{}c@{}} {2, 4}\\{2, 4} \end{tabular}&\begin{tabular}[c]{@{}c@{}} \{1,5,10\}-shot\\\{1,5,10\}-shot \end{tabular}& AC & \ding{55} \\ \hline
 \cite{narwariya2020metab} & Arrhythmia & {ECG} & Metric + Optimization & Supervised, Episodic & Tasks &\begin{tabular}[c]{@{}c@{}} UCR ECG200 \cite{dau2019ucr}\\UCR ECG5000 \cite{dau2019ucr}\\UCR ECGFiveDays \cite{dau2019ucr} \end{tabular}&\begin{tabular}[c]{@{}c@{}} 2\\5\\2 \end{tabular}&\begin{tabular}[c]{@{}c@{}} \{2,5,10,20\}-shot\\\{2,5,10,20\}-shot\\\{2,5,10,20\}-shot \end{tabular}& AC & \ding{55} \\ \hline
 \cite{liu2022side} & Tinnitus & {EEG} & Metric + Optimization & Semi-Supervised, Episodic & Dataset &\begin{tabular}[c]{@{}c@{}} \cite{schaette2011tinnitus}\\\cite{guest2017tinnitus} \end{tabular}& 2 & 2 per subject & F1, AC & \ding{55} \\ \hline
\bottomrule
\end{tabular}%
}
\raggedright
\begin{tablenotes}
All tasks, except modality translation, are classification tasks. Abbreviations: Electrocardiogram (ECG), Bio-Impedance (Bio-Z), Electroencephalogram (EEG), Percentage Root Mean Square Difference (PRD), Weighted Diagnostic Distortion (WDD), Pearson Correlation (PC), Accuracy (AC), Recall (RE), Specificity (SP), G-Mean (GM), True Negative Rate (TNR), True Positive Rate (TPR)
\end{tablenotes}
\end{table*}
Hybrid methods to be presented in this section combine the aforementioned ideas to further improve few-shot learning performance. A popular choice is to train metric-based models through meta-learning optimization algorithms. Li et al. \cite{li2023novel} combined the idea of \ac{MAML} with prototypical networks. Treating each task of \ac{MAML} as a dataset from a specific subject, the authors use \ac{MAML} to pre-train a backbone network and calculate prototypes for each class using the support set. Then, the prototypes guide the selection and pseudo-labeling process of unlabeled target subject data, to provide additional samples for fine-tuning and adapting to the target subject. Experiments were conducted in three \ac{BCI} applications and demonstrated improvements over the baseline models and \ac{MAML} itself. The authors conclude that the meta-learned class centers are useful for guiding the artificial labeling and selection of unlabeled data for improved semi-supervised adaption to target subjects. Narwariya et al. \cite{narwariya2020metab} propose to train a matching network with the reptile meta-learning framework. The proposed framework combines the Reptile meta-learning approach with the triplet loss function to pre-train a residual neural network. Each task of the reptile is treated as a dataset from various domains, such as healthcare and activity recognition. The main advantage of matching networks and metric-based methods is that it is not restrained to the classes they are pre-trained on, and Reptile helps to pre-train a matching network that is easy to fine-tune to new tasks with new classes. Therefore, the reptile-based cross-domain meta-learning enables the pre-trained model to quickly adapt to unseen domains, as well as adapt to tasks with varying numbers of classes without introducing additional class parameters.
\par
Along the same lines of combining metric-based methods and meta-learning, Liu et al. \cite{liu2021few} took inspiration from \cite{triantafillou2019meta} to combine prototypical networks with transfer learning. Triantafillou et al. \cite{triantafillou2019meta} observed that a prototypical network with Euclidean distance measure can be rewritten as a linear classifier with respect to the query embedding:
\begin{equation*}
\resizebox{\columnwidth}{!}{
$\displaystyle
\begin{aligned} 
-\lVert f_\theta(\mathbf{x}_{q,j}) - \mathbf{c}_k \rVert ^2 &= - f_\theta(\mathbf{x}_{q,j})^T f_\theta(\mathbf{x}_{q,j}) + 2\mathbf{c}_k^T f_\theta(\mathbf{x}_{q,j}) - \mathbf{c}_k^T \mathbf{c}_k \\ \notag
    &= 2\mathbf{c}_k^T f_\theta(\mathbf{x}_{q,j}) - \lVert \mathbf{c}_k\rVert^2_2 + constant
\end{aligned}
$}
\end{equation*}
where $constant$ is a scalar that does not affect the prediction, $f_\theta(\mathbf{x}_{q,j})$ is the embedding of the query sample $\mathbf{x}_{q,j}$, and $\lVert \mathbf{c}_k\rVert_2$ is the Euclidean norm of the prototype of class $k$. Therefore, the prototypical network classifier can be rewritten as a linear layer as a function of $f_\theta(\mathbf{x}_{q,j})$, with the $k$-th neuron having weights $\mathbf{W}_k = 2  \mathbf{c}_k$ and biases $\mathbf{b}_k = - \lVert \mathbf{c}_k \rVert^2$. Liu et al. pre-trained the embedding network of the prototypical network using the source dataset, followed by initializing the linear layer using the pre-trained class prototypes for fine-tuning. The proposed paradigm was found to outperform existing few-shot learning baselines, including prototypical networks, matching networks, relation networks, and \ac{MAML}. In a different study, Liu et al. \cite{liu2022side} chose to apply transfer learning to Siamese networks. The proposed Siamese autoencoder is first trained in a semi-supervised manner, through a combination of \ac{MSE} for signal reconstruction and cross-entropy for two individual classifiers: side predictor (left or right ear) and class predictor (normal or tinnitus). The pre-trained encoder is then fine-tuned in an unsupervised manner with the side predictor classifier. The advantage of this framework is the automatic calibration process that does not require annotation of few-shot samples, as fine-tuning involves the use of information from the left/right ear, which can be obtained directly from data collection. 
With a different stream of thought, Akbari et al. \cite{akbari2021meta} combine the idea of model-based methods and meta-learning. The authors propose an encoder-decoder network, where the encoder is shared between subjects and the decoder is personalized. While the encoder weights are trained through backpropagation, the decoder weights are generated by a weight calibration network. The weight calibration network is a meta-learning module that receives the support set as input and generates the weights and biases for a linear layer. The authors claim that with limited samples, the weight calibration network can update the weights of the decoder better than backpropagation-based fine-tuning, which tends to overfit without sufficient data. Furthermore, personalization with a weight-calibration network only requires a forward pass to update the decoder weights, which is particularly appealing for low-power edge devices. However, the weight calibration network is unstable, in the sense that the weight generation process is sensitive to any small perturbations of the support set. Thus, it is paramount to pre-train a weight calibration network that can generalize well across different distributions of support sets. 
\par
As discussed in Section \ref{sec:datasummary}, augmenting the dataset with synthetic samples using data-based methods alone is insufficient to overcome the unique challenges associated with biomedical time series, such as inter-patient variability. Thus, a few works take a step further and design more complex networks to complement data augmentation. For example, Aldahr et al. \cite{aldahr2022addressinga} propose to augment Siamese networks with \ac{GAN} data augmentation. The framework consists of three components: diversity-enhanced data augmentation with \ac{GAN}, temporal-aware feature extraction with graph theory, and Siamese network. The \ac{GAN} model is enhanced with a diversity factor to avoid generating redundant samples by performing a diversity analysis between synthetic \& real samples and between synthetic samples \cite{pascual2019synthetic, farahani2019application}. The graph theory method transforms \ac{EEG} samples into complex weighted graphs, which capture intrinsic hidden patterns and information with minimal information loss. Finally, the Siamese network learns inter-patient seizure patterns through pairs of similar and dissimilar samples from different subjects. Experimental results show that the proposed framework addresses data scarcity and inter-patient variability problems of EEG for seizure detection, outperforming existing methods with less data. Similarly, Kim et al. \cite{kim2022automatic} combined the data augmentation technique SMOTE with a custom-designed model consisting of ResNet and bi-directional \ac{LSTM}. SMOTE was applied to address the problem of class imbalance during pre-training by generating synthetic samples via interpolation between samples of minority classes. An ablation study demonstrates the usefulness of each module when trained on the full dataset. Compared to ShallowConvNet \cite{schirrmeister2017deep} and DeepConvNet \cite{volker2018deep}, the proposed model achieves better few-shot adaptation for cardiac arrhythmia classification datasets, but the results were not desirable for clinical application.
\par
While the aforementioned works have been restricted to the task of classification, Moon et al. \cite{moon2022explainable} explored few-shot open set recognition through a prototypical network with \ac{SVM} classifier. Open set recognition differs from traditional closed set recognition problems, where test samples belong to the same training classes for closed set recognition problems and test samples can contain different classes than the training dataset. To do so, the authors propose a two-stage network. First, an encoder-decoder is trained in a self-supervised manner using triplet and reconstructed prototype loss to minimize the feature embedding distance of inputs from the same subject and maximize the distance of inputs from different subjects. Second, the pre-trained encoder is frozen and few-shot labeled samples are forward passed to construct support vectors for each subject and a decision boundary for a one-class \ac{SVM} is extracted for test predictions. If the embedding of the query sample falls outside the decision boundary of all training classes, then it is considered unrecognized.
\par
\subsubsection{Summary}
Hybrid methods aim to synergize strengths or complement weaknesses of individual methods. They provide promising performance gains compared to existing few-shot learning baselines. This also increases the complexity of the system, which reduces interpretability and the ease of error attribution.

\section{Discussion} \label{sec:future}
\subsection{Challenges and Potential Solutions}
\subsubsection{Lack of Benchmarks and Standardized Evaluation Metrics}
While many few-shot learning methods have been proposed for biomedical time series, it is rather difficult to compare them. A major obstacle to fair comparison is that proposed methods address different clinical applications using different datasets that are often privately collected. Furthermore, most works compare the proposed method against baseline networks that are not designed for few-shot learning. Although it is clear that few-shot learning methods outperform baseline methods not designed for learning from a few samples, it is unclear which few-shot learning architecture performs better. The differences in experimental setups further prevent fair comparison amongst methods. Some methods involve pre-training on external data sources while others do not. Furthermore, while most work report performance using the $N$-way-$K$-shot paradigm, $N$ and $K$ differ significantly across applications and datasets and can even be different when using the same dataset. Furthermore, the $N$-way-$K$-shot support and query sets are usually artificially sampled from a larger dataset. Due to the small sample size, the evaluation results can differ significantly depending on which samples are drawn. When support and query sets are artificially generated, it is important to repeat the sampling for multiple trials and provide an expectation. This calls for a need to develop standardized few-shot learning datasets and benchmarks that comprehensively capture the performance of existing methods to provide a fair and exhaustive comparison across methods.

\subsubsection{How to Choose Pre-Training Dataset}
For some types of few-shot learning methods, such as optimization-based methods, there is a heavy reliance on pre-training the model with a pre-training dataset. However, there is limited research on how the choice of pre-training datasets impacts knowledge transfer and generalization to the target few-shot task. Often, it is preferred to pre-train using datasets similar to the problem or application at hand, but access to such datasets can be difficult, especially if it is an emerging application or niche field. Given the availability of diverse physiological datasets from databases like PhysioNet \cite{PhysioNet}, perhaps it would be preferable to pre-train using a range of datasets across different applications to help the few-shot learning method learn generalizable knowledge. Guidelines on choosing datasets for pre-training could offer insights into factors like the impact of task similarity and the number of pre-training tasks on few-shot knowledge transfer, providing direction for practitioners navigating the complexities of implementing few-shot learning algorithms.

\subsection{Emerging and Promising Fields of Few-Shot Learning}
\subsubsection{LLM}
As the amount of training data and the size of \acp{LLM} continue to increase, \acp{LLM} demonstrate incredible generalization abilities, along with other emergent capabilities. A surprising example is in-context learning, where the model is asked to perform a previously unseen task based on demonstrations as part of the input prompt. \acp{LLM} are shown to be capable of identifying patterns from the prompt and using the next token prediction to generalize to unseen \ac{NLP} tasks \cite{brown2020language}. More recently, similar ideas were adopted for time series data. By converting time series into a text representation, \acp{LLM} can achieve a forecasting performance similar to or exceeding that of purpose-built forecasting methods \cite{jin2023time, gruver2023large}. With recent meta-learning attempts to design better prompts for few-shot in-context learning \cite{hou2022metaprompting}, the idea can also be embraced to optimize conversion from time series to text representation and prompts. Leveraging the powerful capabilities of \acp{LLM} presents many exciting research directions for few-shot learning in the time series domain.

\subsubsection{Multi-Modality}
The vast majority of current methods focus on a singular modality, with a limited number of methods that target multimodal data for human activity recognition. Rapid advancements in wearable technologies \cite{sempionatto2021epidermal, xu2024physicochemical} provide a convenient platform to improve the accessibility of multimodal data sets in a variety of applications. Research on how to effectively fuse multi-modal information and how to learn with unaligned data pave the way for more robust few-shot learning systems that take advantage of complementary information sources. In addition to combining different time series sensors (e.g. \ac{EEG} and \ac{VOG}), multi-modality can also combine different data modalities (e.g. time series and text) to complement each other. A notable example is CaFo \cite{zhang2023prompt}, a Cascade of Foundation models, that leverages diverse prior knowledge of varying pre-trained foundation models to improve few-shot learning. The authors unify the domain knowledge of CLIP \cite{radford2021learning}, DINO \cite{caron2021emerging}, DALL-E \cite{ramesh2021zero} \& GPT-3 \cite{brown2020language} through a "Prompt, Generate, then Cache" pipeline and demonstrated state-of-the-art few-shot learning performance on 11 datasets. Intuitively, triangulating across data modalities encourages the model to learn more discriminative and informative representations, encouraging collaboration of pre-trained knowledge from different modalities to improve few-shot learning performance.


\subsubsection{Neural Adaptive Processes}
While existing few-shot learning methods for biomedical time series focus on adapting model parameters through gradient computations, adapting network parameters using a weight adaptation network has received increasing interest \cite{requeima2019fast}. CNAPs introduced the idea of using an adaptation network that modulates classifier parameters using the support set as input. The idea of hypernetworks, networks that generate weights for classifiers, has also shown promising results for few-shot learning \cite{sendera2023hypershot}. Since backpropagation and gradient computations are not required for adaptation to new tasks, such approaches are computationally efficient, paving the way for personalized continual learning models on wearable devices.

\section{Conclusion}
This paper provides an introduction to the concepts of few-shot learning followed by a comprehensive review of different learning methods for biomedical time series applications. The methodologies were categorized based on how the few-shot learning algorithms modify traditional deep learning training pipelines to overcome the shortage of labeled data, namely data-based, model-based, metric-based, optimization-based, and hybrid methods. The clinical advantages and disadvantages of each method were discussed in relation to traditional deep learning methods. Visual illustrations and tables summarize the key characteristics of concepts and relevant works, presenting concise information for practitioners to choose the most appropriate few-shot learning methods for their custom applications. Trends, challenges, and potential directions for future works were discussed, highlighting the exciting areas and opportunities for the research community.

\renewcommand*{\bibfont}{\footnotesize}
\printbibliography

@article{ahn2023automatic,
  title = {Automatic Stridor Detection Using Small Training Set via Patch-Wise Few-Shot Learning for Diagnosis of Multiple System Atrophy},
  author = {Ahn, Jong Hyeon and Lee, Ju Hwan and Lim, Chae Yeon and Joo, Eun Yeon and Youn, Jinyoung and Chung, Myung Jin and Cho, Jin Whan and Kim, Kyungsu},
  year = {2023},
  journal = {Scientific Reports},
  volume = {13},
  number = {1},
  pages = {10899},
  doi = {10.1038/s41598-023-37620-0},
  pmcid = {PMC10323004},
  pmid = {37407621}
}

@article{akbari2021meta,
  title = {A Meta-Learning Approach for Fast Personalization of Modality Translation Models in Wearable Physiological Sensing},
  author = {Akbari, Ali and Martinez, Jonathan and Jafari, Roozbeh},
  year = {2021},
  journal = {IEEE journal of biomedical and health informatics},
  volume = {26},
  number = {4},
  pages = {1516--1527}
}

@inproceedings{aldahr2022addressinga,
  title = {Addressing {{Inter-Patient Variability}} in {{EEG}}: {{Diversity-Enhanced Data Augmentation}} and {{Few-Shot Learning-based Epilepsy Detection}}},
  booktitle = {2022 {{International Conference}} on {{Healthcare Engineering}} ({{ICHE}})},
  author = {Aldahr, Raghdah Saem and Alanazi, Munid and Ilyas, Mohammad},
  year = {2022},
  pages = {1--7}
}

@article{bhaskarpandit2023how,
  title = {How {{Much Data}} Is {{Enough}}? {{Benchmarking Transfer Learning}} for {{Few Shot ECG Image Classification}}},
  author = {Bhaskarpandit, Sathvik},
  year = {2023}
}

@article{bhosale2022calibration,
  title = {Calibration Free Meta Learning Based Approach for Subject Independent {{EEG}} Emotion Recognition},
  author = {Bhosale, Swapnil and Chakraborty, Rupayan and Kopparapu, Sunil Kumar},
  year = {2022},
  journal = {Biomedical Signal Processing and Control},
  volume = {72},
  pages = {103289}
}

@article{burrello2019hyperdimensionala,
  title = {Hyperdimensional Computing with Local Binary Patterns: {{One-shot}} Learning of Seizure Onset and Identification of Ictogenic Brain Regions Using Short-Time {{iEEG}} Recordings},
  author = {Burrello, Alessio and Schindler, Kaspar and Benini, Luca and Rahimi, Abbas},
  year = {2019},
  journal = {IEEE Transactions on Biomedical Engineering},
  volume = {67},
  number = {2},
  pages = {601--613}
}

@article{compagnon2020learninga,
  title = {Learning Personalized {{ADL}} Recognition Models from Few Raw Data},
  author = {Compagnon, Paul and Lefebvre, Gr{\'e}goire and Duffner, Stefan and Garcia, Christophe},
  year = {2020},
  journal = {Artificial Intelligence in Medicine},
  volume = {107},
  pages = {101916},
  doi = {10.1016/j.artmed.2020.101916},
  pmid = {32828455}
}

@article{duan2020zero,
  title = {Zero-Shot Learning for {{EEG}} Classification in Motor Imagery-Based {{BCI}} System},
  author = {Duan, Lili and Li, Jie and Ji, Hongfei and Pang, Zilong and Zheng, Xuanci and Lu, Rongrong and Li, Maozhen and Zhuang, Jie},
  year = {2020},
  journal = {IEEE Transactions on Neural Systems and Rehabilitation Engineering},
  volume = {28},
  number = {11},
  pages = {2411--2419}
}

@article{elnaggar2023sleepa,
  title = {Sleep Posture One-Shot Learning Framework Based on Extremity Joint Kinematics: {{In-silico}} and in-Vivo Case Studies},
  author = {Elnaggar, Omar and Coenen, Frans and Hopkinson, Andrew and Mason, Lyndon and Paoletti, Paolo},
  year = {2023},
  journal = {Information Fusion},
  volume = {95},
  pages = {215--236}
}

@inproceedings{gupta2021similarityb,
  title = {Similarity {{Learning}} Based {{Few Shot Learning}} for {{ECG Time Series Classification}}},
  booktitle = {2021 {{Digital Image Computing}}: {{Techniques}} and {{Applications}} ({{DICTA}})},
  author = {Gupta, Priyanka and Bhaskarpandit, Sathvik and Gupta, Manik},
  year = {2021},
  pages = {1--8}
}

@incollection{hartmann2023high,
  title = {High-{{Level Features}} for {{Human Activity Recognition}} and {{Modeling}}},
  booktitle = {Biomedical {{Engineering Systems}} and {{Technologies}}},
  author = {Hartmann, Yale and Liu, Hui and Schultz, Tanja},
  editor = {Roque, Ana Cec{\'i}lia A. and Gracanin, Denis and Lorenz, Ronny and Tsanas, Athanasios and Bier, Nathalie and Fred, Ana and Gamboa, Hugo},
  year = {2023},
  volume = {1814},
  pages = {141--163},
  doi = {10.1007/978-3-031-38854-5_8}
}

@article{hernandez2023prototypical,
  title = {A Prototypical Network for Few-Shot Recognition of Speech Imagery Data},
  author = {{Hernandez-Galvan}, Alan and {Ramirez-Alonso}, Graciela and {Ramirez-Quintana}, Juan},
  year = {2023},
  journal = {Biomedical Signal Processing and Control},
  volume = {86},
  pages = {105154}
}

@article{hur2023genhpf,
  title = {{{GenHPF}}: {{General Healthcare Predictive Framework}} for {{Multi-task Multi-source Learning}}},
  author = {Hur, Kyunghoon and Oh, Jungwoo and Kim, Junu and Kim, Jiyoun and Lee, Min Jae and Cho, Eunbyeol and Moon, Seong-Eun and Kim, Young-Hak and Atallah, Louis and Choi, Edward},
  year = {2023},
  journal = {IEEE Journal of Biomedical and Health Informatics}
}

@inproceedings{hwang2019ezsl,
  title = {{{EZSL-GAN}}: {{EEG-based}} Zero-Shot Learning Approach Using a Generative Adversarial Network},
  booktitle = {2019 7th {{International Winter Conference}} on {{Brain-Computer Interface}} ({{BCI}})},
  author = {Hwang, Sunhee and Hong, Kibeom and Son, Guiyoung and Byun, Hyeran},
  year = {2019},
  pages = {1--4}
}

@article{kim2022automatic,
  title = {Automatic Cardiac Arrhythmia Classification Using Residual Network Combined with Long Short-Term Memory},
  author = {Kim, Yun Kwan and Lee, Minji and Song, Hee Seok and Lee, Seong-Whan},
  year = {2022},
  journal = {IEEE Transactions on Instrumentation and Measurement},
  volume = {71},
  pages = {1--17}
}

@inproceedings{lan2020gazegraph,
  title = {{{GazeGraph}}: Graph-Based Few-Shot Cognitive Context Sensing from Human Visual Behavior},
  booktitle = {Proceedings of the 18th {{Conference}} on {{Embedded Networked Sensor Systems}}},
  author = {Lan, Guohao and Heit, Bailey and Scargill, Tim and Gorlatova, Maria},
  year = {2020},
  pages = {422--435},
  doi = {10.1145/3384419.3430774}
}

@inproceedings{lee2023source,
  title = {Source-Free {{Subject Adaptation}} for {{EEG-based Visual Recognition}}},
  booktitle = {2023 11th {{International Winter Conference}} on {{Brain-Computer Interface}} ({{BCI}})},
  author = {Lee, Pilhyeon and Jeon, Seogkyu and Hwang, Sunhee and Shin, Minjung and Byun, Hyeran},
  year = {2023},
  pages = {1--6}
}

@article{li2021oneb,
  title = {A {{One-Dimensional Siamese Few-Shot Learning Approach}} for {{ECG Classification}} under {{Limited Data}}},
  author = {Li, Zongjin and Wang, Huan and Liu, Xinwen},
  year = {2021},
  journal = {Annual International Conference of the IEEE Engineering in Medicine and Biology Society. IEEE Engineering in Medicine and Biology Society. Annual International Conference},
  volume = {2021},
  pages = {455--458},
  doi = {10.1109/EMBC46164.2021.9630622},
  pmid = {34891331}
}

@article{li2022meta,
  title = {Meta-{{Learning}} for {{Fast}} and {{Privacy-Preserving Source Knowledge Transfer}} of {{EEG-Based BCIs}}},
  author = {Li, Siyang and Wu, Huanyu and Ding, Lieyun and Wu, Dongrui},
  year = {2022},
  journal = {IEEE Computational Intelligence Magazine},
  volume = {17},
  number = {4},
  pages = {16--26}
}

@article{li2023novel,
  title = {A Novel Semi-Supervised Meta Learning Method for Subject-Transfer Brain\textendash Computer Interface},
  author = {Li, Jingcong and Wang, Fei and Huang, Haiyun and Qi, Feifei and Pan, Jiahui},
  year = {2023},
  journal = {Neural Networks},
  volume = {163},
  pages = {195--204}
}

@article{liu2021few,
  title = {Few-Shot Learning for Cardiac Arrhythmia Detection Based on Electrocardiogram Data from Wearable Devices},
  author = {Liu, Tianyu and Yang, Yukang and Fan, Wenhui and Wu, Cheng},
  year = {2021},
  journal = {Digital Signal Processing},
  volume = {116},
  pages = {103094}
}

@article{liu2022side,
  title = {Side-Aware Meta-Learning for Cross-Dataset Listener Diagnosis with Subjective Tinnitus},
  author = {Liu, Zhe and Li, Yun and Yao, Lina and Lucas, Molly and Monaghan, Jessica JM and Zhang, Yu},
  year = {2022},
  journal = {IEEE Transactions on Neural Systems and Rehabilitation Engineering},
  volume = {30},
  pages = {2352--2361}
}

@article{lv2022exploratory,
  title = {An Exploratory Study of Transfer Learning Frameworks in the Context of Few Available Shots of Neurophysiological Signals},
  author = {Lv, Yizhi and Xue, Jianing and Duan, Feng and Sun, Zhe and Li, Junhua},
  year = {2022},
  journal = {Computers and Electrical Engineering},
  volume = {101},
  pages = {108091}
}

@incollection{ma2023fewa,
  title = {Few-{{Shot Class-Incremental Learning}} for {{EEG-Based Emotion Recognition}}},
  booktitle = {Neural {{Information Processing}}},
  author = {Ma, Tian-Fang and Zheng, Wei-Long and Lu, Bao-Liang},
  editor = {Tanveer, Mohammad and Agarwal, Sonali and Ozawa, Seiichi and Ekbal, Asif and Jatowt, Adam},
  year = {2023},
  volume = {1792},
  pages = {445--455},
  doi = {10.1007/978-981-99-1642-9_38}
}

@article{mccartney2019zeroa,
  title = {A Zero-Shot Learning Approach to the Development of Brain-Computer Interfaces for Image Retrieval},
  author = {McCartney, Ben and {Martinez-del-Rincon}, Jesus and Devereux, Barry and Murphy, Brian},
  year = {2019},
  journal = {Plos one},
  volume = {14},
  number = {9},
  pages = {e0214342}
}

@article{meyer2022u,
  title = {U-{{HAR}}: {{A Convolutional Approach}} to {{Human Activity Recognition Combining Head}} and {{Eye Movements}} for {{Context-Aware Smart Glasses}}},
  author = {Meyer, Johannes and Frank, Adrian and Schlebusch, Thomas and Kasneci, Enkelejda},
  year = {2022},
  journal = {Proceedings of the ACM on Human-Computer Interaction},
  volume = {6},
  number = {ETRA},
  pages = {1--19},
  doi = {10.1145/3530884}
}

@article{moon2022explainable,
  title = {Explainable Gait Recognition with Prototyping Encoder\textendash Decoder},
  author = {Moon, Jucheol and Shin, Yong-Min and Park, Jin-Duk and Minaya, Nelson Hebert and Shin, Won-Yong and Choi, Sang-Il},
  year = {2022},
  journal = {Plos one},
  volume = {17},
  number = {3},
  pages = {e0264783}
}

@inproceedings{munia2021imbalanceda,
  title = {Imbalanced Eeg Analysis Using One-Shot Learning with Siamese Neural Network},
  booktitle = {2021 {{IEEE}} 9th {{International Conference}} on {{Healthcare Informatics}} ({{ICHI}})},
  author = {Munia, Munawara Saiyara and Hosseini, Seyyed MohammadSaleh and Nourani, Mehrdad and Harvey, Jay and Dave, Hina},
  year = {2021},
  pages = {4--12}
}

@inproceedings{narwariya2020metab,
  title = {Meta-{{Learning}} for {{Few-Shot Time Series Classification}}},
  booktitle = {Proceedings of the 7th {{ACM IKDD CoDS}} and 25th {{COMAD}}},
  author = {Narwariya, Jyoti and Malhotra, Pankaj and Vig, Lovekesh and Shroff, Gautam and Vishnu, T. V.},
  year = {2020},
  pages = {28--36},
  doi = {10.1145/3371158.3371162}
}

@article{nazari2022epilepsya,
  title = {Epilepsy Seizure Prediction with Few-Shot Learning Method},
  author = {Nazari, Jamal and Motie Nasrabadi, Ali and Menhaj, Mohammad Bagher and Raiesdana, Somayeh},
  year = {2022},
  journal = {Brain Informatics},
  volume = {9},
  number = {1},
  pages = {21},
  doi = {10.1186/s40708-022-00170-8}
}

@article{ng2023few,
  title = {Few-Shot Transfer Learning for Personalized Atrial Fibrillation Detection Using Patient-Based Siamese Network with Single-Lead {{ECG}} Records},
  author = {Ng, Yiuwai and Liao, Min-Tsun and Chen, Ting-Li and Lee, Chih-Kuo and Chou, Cheng-Ying and Wang, Weichung},
  year = {2023},
  journal = {Artificial Intelligence in Medicine},
  volume = {144},
  pages = {102644}
}

@inproceedings{pan2022few,
  title = {Few-Shot {{Egocentric Multimodal Activity Recognition}}},
  booktitle = {{{ACM Multimedia Asia}}},
  author = {Pan, Jinxing and Yang, Xiaoshan and Huang, Yi and Xu, Changsheng},
  year = {2022},
  pages = {1--7},
  doi = {10.1145/3469877.3490603}
}

@article{pan2023multiplea,
  title = {Multiple {{Scale Convolutional Few Shot Learning Networks}} for {{Online P300-based Brain-Computer Interface}} and {{Its Application}} to {{Patients}} with {{Disorder}} of {{Consciousness}}},
  author = {Pan, Jiahui and Cai, Honghua and Huang, Haiyun and He, Yanbin and Li, Yuanqing},
  year = {2023},
  journal = {IEEE Transactions on Instrumentation and Measurement}
}

@article{phunruangsakao2022deep,
  title = {Deep Adversarial Domain Adaptation with Few-Shot Learning for Motor-Imagery Brain-Computer Interface},
  author = {Phunruangsakao, Chatrin and Achanccaray, David and Hayashibe, Mitsuhiro},
  year = {2022},
  journal = {IEEE Access},
  volume = {10},
  pages = {57255--57265}
}

@article{poulain2022fewc,
  title = {Few-{{Shot Learning}} with {{Semi-Supervised Transformers}} for {{Electronic Health Records}}},
  author = {Poulain, Raphael and Gupta, Mehak and Beheshti, Rahmatollah},
  year = {2022},
  journal = {Proceedings of Machine Learning Research},
  volume = {182},
  pages = {853--873},
  pmcid = {PMC10399128},
  pmid = {37538125}
}

@inproceedings{rahimian2021fewa,
  title = {Few-Shot Learning for Decoding Surface Electromyography for Hand Gesture Recognition},
  booktitle = {{{ICASSP}} 2021-2021 {{IEEE International Conference}} on {{Acoustics}}, {{Speech}} and {{Signal Processing}} ({{ICASSP}})},
  author = {Rahimian, Elahe and Zabihi, Soheil and Asif, Amir and Atashzar, S. Farokh and Mohammadi, Arash},
  year = {2021},
  pages = {1300--1304}
}

@article{saeed2021sense,
  title = {Sense and Learn: {{Self-supervision}} for Omnipresent Sensors},
  author = {Saeed, Aaqib and Ungureanu, Victor and Gfeller, Beat},
  year = {2021},
  journal = {Machine Learning with Applications},
  volume = {6},
  pages = {100152}
}

@inproceedings{salekin2021understanding,
  title = {Understanding Autism: The Power of {{EEG}} Harnessed by Prototypical Learning},
  booktitle = {Proceedings of the {{Workshop}} on {{Medical Cyber Physical Systems}} and {{Internet}} of {{Medical Things}}},
  author = {Salekin, Asif and Russo, Natalie},
  year = {2021},
  pages = {12--16},
  doi = {10.1145/3446913.3460317}
}

@article{soroushmojdehi2022transfer,
  title = {Transfer Learning in Hand Movement Intention Detection Based on Surface Electromyography Signals},
  author = {Soroushmojdehi, Rahil and Javadzadeh, Sina and Pedrocchi, Alessandra and Gandolla, Marta},
  year = {2022},
  journal = {Frontiers in Neuroscience},
  volume = {16},
  pages = {977328}
}

@inproceedings{suo2020tadanet,
  title = {{{TAdaNet}}: {{Task-Adaptive Network}} for {{Graph-Enriched Meta-Learning}}},
  booktitle = {Proceedings of the 26th {{ACM SIGKDD International Conference}} on {{Knowledge Discovery}} \& {{Data Mining}}},
  author = {Suo, Qiuling and Chou, Jingyuan and Zhong, Weida and Zhang, Aidong},
  year = {2020},
  pages = {1789--1799},
  doi = {10.1145/3394486.3403230}
}

@inproceedings{suwannaphong2022radioa,
  title = {Radio Signal Strength Indication Augmentation for One-Shot Learning in Indoor Localisation},
  booktitle = {Proceedings of the 1st {{ACM Workshop}} on {{Smart Wearable Systems}} and {{Applications}}},
  author = {Suwannaphong, Thanaphon and McConville, Ryan and Craddock, Ian},
  year = {2022},
  pages = {7--12},
  doi = {10.1145/3556560.3560714}
}

@inproceedings{tam2022siamese,
  title = {Siamese {{Convolutional Neural Network}} and {{Few-Shot Learning}} for {{Embedded Gesture Recognition}}},
  booktitle = {2022 20th {{IEEE Interregional NEWCAS Conference}} ({{NEWCAS}})},
  author = {Tam, Simon and Boukadoum, Mounir and {Campeau-Lecours}, Alexandre and Gosselin, Benoit},
  year = {2022},
  pages = {114--118}
}

@inproceedings{tang2020interpretable,
  title = {Interpretable Time-Series Classification on Few-Shot Samples},
  booktitle = {2020 {{International Joint Conference}} on {{Neural Networks}} ({{IJCNN}})},
  author = {Tang, Wensi and Liu, Lu and Long, Guodong},
  year = {2020},
  pages = {1--8}
}

@article{wang2023generalized,
  title = {A {{Generalized Zero-Shot Learning Scheme}} for {{SSVEP-Based BCI System}}},
  author = {Wang, Xietian and Liu, Aiping and Wu, Le and Li, Chang and Liu, Yu and Chen, Xun},
  year = {2023},
  journal = {IEEE Transactions on Neural Systems and Rehabilitation Engineering},
  volume = {31},
  pages = {863--874}
}

@article{wang2023personalized,
  title = {Personalized {{Modeling}} of {{Blood Pressure}} with {{Photoplethysmography}}: An {{Error-Feedback Incremental Support Vector Regression Model}}},
  author = {Wang, Dingliang and Yang, Xuezhi and Wu, Jun and Wang, Wenjin},
  year = {2023},
  journal = {IEEE Internet of Things Journal}
}

@article{wang2023similarity,
  title = {Similarity Function for One-Shot Learning to Enhance the Flexibility of Myoelectric Interfaces},
  author = {Wang, Xiang and Zhang, Xu and Chen, Xiang and Chen, Xun and Lv, Zhao and Liang, Zhen},
  year = {2023},
  journal = {IEEE Transactions on Neural Systems and Rehabilitation Engineering},
  volume = {31},
  pages = {1697--1706}
}

@inproceedings{wu2022does,
  title = {Does {{Meta-Learning Improve EEG Motor Imagery Classification}}?},
  booktitle = {2022 44th {{Annual International Conference}} of the {{IEEE Engineering}} in {{Medicine}} \& {{Biology Society}} ({{EMBC}})},
  author = {Wu, Xiaoli and Chan, Rosa HM},
  year = {2022},
  pages = {4048--4051}
}

@article{you2022few,
  title = {A {{Few-Shot Learning-Based EEG}} and {{Stage Transition Sequence Generator}} for {{Improving Sleep Staging Performance}}},
  author = {You, Yuyang and Guo, Xiaoyu and Zhong, Xuyang and Yang, Zhihong},
  year = {2022},
  journal = {Biomedicines},
  volume = {10},
  number = {12},
  pages = {3006},
  doi = {10.3390/biomedicines10123006},
  pmcid = {PMC9775526},
  pmid = {36551762}
}

@inproceedings{youssef2022model,
  title = {Model {{Personalization}} with {{Static}} and {{Dynamic Patients}}' {{Data}}},
  booktitle = {2022 {{IEEE International Conference}} on {{Data Mining Workshops}} ({{ICDMW}})},
  author = {Youssef, Paul and Schl{\"o}tterer, J{\"o}rg and Imangaliyev, Sultan and Seifert, Christin},
  year = {2022},
  pages = {324--333}
}

@inproceedings{zhang2019metapred,
  title = {{{MetaPred}}: {{Meta-Learning}} for {{Clinical Risk Prediction}} with {{Limited Patient Electronic Health Records}}},
  booktitle = {Proceedings of the 25th {{ACM SIGKDD International Conference}} on {{Knowledge Discovery}} \& {{Data Mining}}},
  author = {Zhang, Xi Sheryl and Tang, Fengyi and Dodge, Hiroko H. and Zhou, Jiayu and Wang, Fei},
  year = {2019},
  pages = {2487--2495},
  doi = {10.1145/3292500.3330779}
}

@article{zhang2022few,
  title = {Few-Shot Learning for Fine-Grained Emotion Recognition Using Physiological Signals},
  author = {Zhang, Tianyi and El Ali, Abdallah and Hanjalic, Alan and Cesar, Pablo},
  year = {2022},
  journal = {IEEE Transactions on Multimedia}
}

@article{zhang2022patienta,
  title = {A Patient-Specific Closed-Loop Epilepsy Management {{SoC}} with One-Shot Learning and Online Tuning},
  author = {Zhang, Miaolin and Zhang, Lian and Tsai, Chne-Wuen and Yoo, Jerald},
  year = {2022},
  journal = {IEEE Journal of Solid-State Circuits},
  volume = {57},
  number = {4},
  pages = {1049--1060}
}

@inproceedings{zhu2021unsupervised,
  title = {Unsupervised Domain Adaptation for Cross-Subject Few-Shot Neurological Symptom Detection},
  booktitle = {2021 10th {{International IEEE}}/{{EMBS Conference}} on {{Neural Engineering}} ({{NER}})},
  author = {Zhu, Bingzhao and Shoaran, Mahsa},
  year = {2021},
  pages = {181--184}
}

@misc{ge2022few,
  title = {Few-Shot Learning for Medical Text: {{A}} Systematic Review},
  author = {Ge, Yao and Guo, Yuting and Yang, Yuan-Chi and {Al-Garadi}, Mohammed Ali and Sarker, Abeed},
  year = {2022},
  number = {arXiv:2204.14081},
  doi = {10.48550/arXiv.2204.14081}
}

@article{ge2023few,
  title = {Few-Shot Learning for Medical Text: {{A}} Review of Advances, Trends, and Opportunities},
  author = {Ge, Yao and Guo, Yuting and Das, Sudeshna and {Al-Garadi}, Mohammed Ali and Sarker, Abeed},
  year = {2023},
  journal = {Journal of Biomedical Informatics},
  volume = {144},
  pages = {104458},
  doi = {10.1016/j.jbi.2023.104458}
}

@misc{jadon2023overview,
  title = {An {{Overview}} of {{Deep Learning Architectures}} in {{Few-Shot Learning Domain}}},
  author = {Jadon, Shruti and Jadon, Aryan},
  year = {2023},
  number = {arXiv:2008.06365}
}

@incollection{kotia2021few,
  title = {Few {{Shot Learning}} for {{Medical Imaging}}},
  booktitle = {Machine {{Learning Algorithms}} for {{Industrial Applications}}},
  author = {Kotia, Jai and Kotwal, Adit and Bharti, Rishika and Mangrulkar, Ramchandra},
  editor = {Das, Santosh Kumar and Das, Shom Prasad and Dey, Nilanjan and Hassanien, Aboul-Ella},
  year = {2021},
  pages = {107--132},
  doi = {10.1007/978-3-030-50641-4_7}
}

@misc{pachetti2023systematic,
  title = {A {{Systematic Review}} of {{Few-Shot Learning}} in {{Medical Imaging}}},
  author = {Pachetti, Eva and Colantonio, Sara},
  year = {2023},
  number = {arXiv:2309.11433}
}

@misc{parnami2022learning,
  title = {Learning from {{Few Examples}}: {{A Summary}} of {{Approaches}} to {{Few-Shot Learning}}},
  author = {Parnami, Archit and Lee, Minwoo},
  year = {2022},
  number = {arXiv:2203.04291}
}

@article{song2023comprehensive,
  title = {A {{Comprehensive Survey}} of {{Few-shot Learning}}: {{Evolution}}, {{Applications}}, {{Challenges}}, and {{Opportunities}}},
  author = {Song, Yisheng and Wang, Ting and Cai, Puyu and Mondal, Subrota K. and Sahoo, Jyoti Prakash},
  year = {2023},
  journal = {ACM Computing Surveys},
  volume = {55},
  number = {13s},
  pages = {271:1--271:40},
  doi = {10.1145/3582688}
}

@article{wang2021generalizing,
  title = {Generalizing from a {{Few Examples}}: {{A Survey}} on {{Few-shot Learning}}},
  author = {Wang, Yaqing and Yao, Quanming and Kwok, James T. and Ni, Lionel M.},
  year = {2021},
  journal = {ACM Computing Surveys},
  volume = {53},
  number = {3},
  pages = {1--34},
  doi = {10.1145/3386252}
}

@article{tsoumplekas2024toward,
  title={Toward green and human-like artificial intelligence: A complete survey on contemporary few-shot learning approaches},
  author={Tsoumplekas, Georgios and Li, Vladislav and Argyriou, Vasileios and Lytos, Anastasios and Fountoukidis, Eleftherios and Goudos, Sotirios K and Moscholios, Ioannis D and Sarigiannidis, Panagiotis},
  journal={arXiv preprint arXiv:2402.03017},
  year={2024}
}

@inproceedings{arjovsky2017wasserstein,
  title={Wasserstein generative adversarial networks},
  author={Arjovsky, Martin and Chintala, Soumith and Bottou, L{\'e}on},
  booktitle={International conference on machine learning},
  pages={214--223},
  year={2017},
  organization={PMLR}
}

@article{goodfellow2014generative,
  title={Generative adversarial nets},
  author={Goodfellow, Ian and Pouget-Abadie, Jean and Mirza, Mehdi and Xu, Bing and Warde-Farley, David and Ozair, Sherjil and Courville, Aaron and Bengio, Yoshua},
  journal={Advances in neural information processing systems},
  volume={27},
  year={2014}
}

@article{mirza2014conditional,
  title={Conditional generative adversarial nets},
  author={Mirza, Mehdi and Osindero, Simon},
  journal={arXiv preprint arXiv:1411.1784},
  year={2014}
}

@article{mikolov2013efficient,
  title={Efficient estimation of word representations in vector space},
  author={Mikolov, Tomas and Chen, Kai and Corrado, Greg and Dean, Jeffrey},
  journal={arXiv preprint arXiv:1301.3781},
  year={2013}
}

@article{bromley1993signature,
  title={Signature verification using a" siamese" time delay neural network},
  author={Bromley, Jane and Guyon, Isabelle and LeCun, Yann and S{\"a}ckinger, Eduard and Shah, Roopak},
  journal={Advances in neural information processing systems},
  volume={6},
  year={1993}
}

@inproceedings{bertinetto2016fully,
  title={Fully-convolutional siamese networks for object tracking},
  author={Bertinetto, Luca and Valmadre, Jack and Henriques, Joao F and Vedaldi, Andrea and Torr, Philip HS},
  booktitle={Computer Vision--ECCV 2016 Workshops: Amsterdam, The Netherlands, October 8-10 and 15-16, 2016, Proceedings, Part II 14},
  pages={850--865},
  year={2016},
  organization={Springer}
}

@inproceedings{mueller2016siamese,
  title={Siamese recurrent architectures for learning sentence similarity},
  author={Mueller, Jonas and Thyagarajan, Aditya},
  booktitle={Proceedings of the AAAI conference on artificial intelligence},
  volume={30},
  number={1},
  year={2016}
}

@inproceedings{bandara2022transformer,
  title={A transformer-based siamese network for change detection},
  author={Bandara, Wele Gedara Chaminda and Patel, Vishal M},
  booktitle={IGARSS 2022-2022 IEEE International Geoscience and Remote Sensing Symposium},
  pages={207--210},
  year={2022},
  organization={IEEE}
}

@article{jin2021multi,
  title={Multi-scale contrastive siamese networks for self-supervised graph representation learning},
  author={Jin, Ming and Zheng, Yizhen and Li, Yuan-Fang and Gong, Chen and Zhou, Chuan and Pan, Shirui},
  journal={arXiv preprint arXiv:2105.05682},
  year={2021}
}

@article{hudgins1993new,
  title={A new strategy for multifunction myoelectric control},
  author={Hudgins, Bernard and Parker, Philip and Scott, Robert N},
  journal={IEEE transactions on biomedical engineering},
  volume={40},
  number={1},
  pages={82--94},
  year={1993},
  publisher={IEEE}
}

@article{sun2022euler,
  title={Euler common spatial patterns for EEG classification},
  author={Sun, Jing and Wei, Mengting and Luo, Ning and Li, Zhanli and Wang, Haixian},
  journal={Medical \& Biological Engineering \& Computing},
  volume={60},
  number={3},
  pages={753--767},
  year={2022},
  publisher={Springer}
}

@inproceedings{wertheimer2021few,
  title={Few-shot classification with feature map reconstruction networks},
  author={Wertheimer, Davis and Tang, Luming and Hariharan, Bharath},
  booktitle={Proceedings of the IEEE/CVF conference on computer vision and pattern recognition},
  pages={8012--8021},
  year={2021}
}

@article{snell2017prototypical,
  title={Prototypical networks for few-shot learning},
  author={Snell, Jake and Swersky, Kevin and Zemel, Richard},
  journal={Advances in neural information processing systems},
  volume={30},
  year={2017}
}

@article{vinyals2016matching,
  title={Matching networks for one shot learning},
  author={Vinyals, Oriol and Blundell, Charles and Lillicrap, Timothy and Wierstra, Daan and others},
  journal={Advances in neural information processing systems},
  volume={29},
  year={2016}
}

@inproceedings{finn2017model,
  title={Model-agnostic meta-learning for fast adaptation of deep networks},
  author={Finn, Chelsea and Abbeel, Pieter and Levine, Sergey},
  booktitle={International conference on machine learning},
  pages={1126--1135},
  year={2017},
  organization={PMLR}
}

@article{nichol2018first,
  title={On first-order meta-learning algorithms},
  author={Nichol, Alex and Achiam, Joshua and Schulman, John},
  journal={arXiv preprint arXiv:1803.02999},
  year={2018}
}

@article{lawhern2018eegnet,
  title={EEGNet: a compact convolutional neural network for EEG-based brain--computer interfaces},
  author={Lawhern, Vernon J and Solon, Amelia J and Waytowich, Nicholas R and Gordon, Stephen M and Hung, Chou P and Lance, Brent J},
  journal={Journal of neural engineering},
  volume={15},
  number={5},
  pages={056013},
  year={2018},
  publisher={iOP Publishing}
}

@article{triantafillou2019meta,
  title={Meta-dataset: A dataset of datasets for learning to learn from few examples},
  author={Triantafillou, Eleni and Zhu, Tyler and Dumoulin, Vincent and Lamblin, Pascal and Evci, Utku and Xu, Kelvin and Goroshin, Ross and Gelada, Carles and Swersky, Kevin and Manzagol, Pierre-Antoine and others},
  journal={arXiv preprint arXiv:1903.03096},
  year={2019}
}

@inproceedings{mcmahan2017communication,
  title={Communication-efficient learning of deep networks from decentralized data},
  author={McMahan, Brendan and Moore, Eider and Ramage, Daniel and Hampson, Seth and y Arcas, Blaise Aguera},
  booktitle={Artificial intelligence and statistics},
  pages={1273--1282},
  year={2017},
  organization={PMLR}
}

@article{pascual2019synthetic,
  title={Synthetic epileptic brain activities using GANs},
  author={Pascual, Dami{\'a}n and Aminifar, Amir and Atienza, David and Ryvlin, Philippe and Wattenhofer, Roger},
  journal={Machine Learning for Health (ML4H) at NeurIPS},
  year={2019}
}

@article{farahani2019application,
  title={Application of graph theory for identifying connectivity patterns in human brain networks: a systematic review},
  author={Farahani, Farzad V and Karwowski, Waldemar and Lighthall, Nichole R},
  journal={frontiers in Neuroscience},
  volume={13},
  pages={585},
  year={2019},
  publisher={Frontiers Media SA}
}

@article{schirrmeister2017deep,
  title={Deep learning with convolutional neural networks for EEG decoding and visualization},
  author={Schirrmeister, Robin Tibor and Springenberg, Jost Tobias and Fiederer, Lukas Dominique Josef and Glasstetter, Martin and Eggensperger, Katharina and Tangermann, Michael and Hutter, Frank and Burgard, Wolfram and Ball, Tonio},
  journal={Human brain mapping},
  volume={38},
  number={11},
  pages={5391--5420},
  year={2017},
  publisher={Wiley Online Library}
}

@inproceedings{volker2018deep,
  title={Deep transfer learning for error decoding from non-invasive EEG},
  author={V{\"o}lker, Martin and Schirrmeister, Robin T and Fiederer, Lukas DJ and Burgard, Wolfram and Ball, Tonio},
  booktitle={2018 6th International Conference on Brain-Computer Interface (BCI)},
  pages={1--6},
  year={2018},
  organization={IEEE}
}

@article{kemp2000analysis,
  title={Analysis of a sleep-dependent neuronal feedback loop: the slow-wave microcontinuity of the EEG},
  author={Kemp, Bob and Zwinderman, Aeilko H and Tuk, Bert and Kamphuisen, Hilbert AC and Oberye, Josefien JL},
  journal={IEEE Transactions on Biomedical Engineering},
  volume={47},
  number={9},
  pages={1185--1194},
  year={2000},
  publisher={IEEE}
}

@inproceedings{spampinato2017deep,
  title={Deep learning human mind for automated visual classification},
  author={Spampinato, Concetto and Palazzo, Simone and Kavasidis, Isaak and Giordano, Daniela and Souly, Nasim and Shah, Mubarak},
  booktitle={Proceedings of the IEEE conference on computer vision and pattern recognition},
  pages={6809--6817},
  year={2017}
}

@article{johnson2020mimic,
  title={Mimic-iv},
  author={Johnson, Alistair and Bulgarelli, Lucas and Pollard, Tom and Horng, Steven and Celi, Leo Anthony and Mark, Roger},
  journal={PhysioNet. Available online at: https://physionet. org/content/mimiciv/1.0/(accessed August 23, 2021)},
  year={2020}
}

@article{walonoski2018synthea,
  title={Synthea: An approach, method, and software mechanism for generating synthetic patients and the synthetic electronic health care record},
  author={Walonoski, Jason and Kramer, Mark and Nichols, Joseph and Quina, Andre and Moesel, Chris and Hall, Dylan and Duffett, Carlton and Dube, Kudakwashe and Gallagher, Thomas and McLachlan, Scott},
  journal={Journal of the American Medical Informatics Association},
  volume={25},
  number={3},
  pages={230--238},
  year={2018},
  publisher={Oxford University Press}
}

@article{all2019all,
  title={The “All of Us” research program},
  author={All of Us Research Program Investigators},
  journal={New England Journal of Medicine},
  volume={381},
  number={7},
  pages={668--676},
  year={2019},
  publisher={Mass Medical Soc}
}

@article{byrne2018residential,
  title={Residential wearable RSSI and accelerometer measurements with detailed location annotations},
  author={Byrne, Dallan and Kozlowski, Michal and Santos-Rodriguez, Raul and Piechocki, Robert and Craddock, Ian},
  journal={Scientific data},
  volume={5},
  number={1},
  pages={1--14},
  year={2018},
  publisher={Nature Publishing Group}
}

@article{liu2020beta,
  title={BETA: A large benchmark database toward SSVEP-BCI application},
  author={Liu, Bingchuan and Huang, Xiaoshan and Wang, Yijun and Chen, Xiaogang and Gao, Xiaorong},
  journal={Frontiers in neuroscience},
  volume={14},
  pages={627},
  year={2020},
  publisher={Frontiers Media SA}
}

@article{wang2016benchmark,
  title={A benchmark dataset for SSVEP-based brain--computer interfaces},
  author={Wang, Yijun and Chen, Xiaogang and Gao, Xiaorong and Gao, Shangkai},
  journal={IEEE Transactions on Neural Systems and Rehabilitation Engineering},
  volume={25},
  number={10},
  pages={1746--1752},
  year={2016},
  publisher={IEEE}
}

@inproceedings{chatzaki2017human,
  title={Human daily activity and fall recognition using a smartphone’s acceleration sensor},
  author={Chatzaki, Charikleia and Pediaditis, Matthew and Vavoulas, George and Tsiknakis, Manolis},
  booktitle={Information and Communication Technologies for Ageing Well and e-Health: Second International Conference, ICT4AWE 2016, Rome, Italy, April 21-22, 2016, Revised Selected Papers 2},
  pages={100--118},
  year={2017},
  organization={Springer}
}

@article{goldberger2000physiobank,
  title={PhysioBank, PhysioToolkit, and PhysioNet: components of a new research resource for complex physiologic signals},
  author={Goldberger, Ary L and Amaral, Luis AN and Glass, Leon and Hausdorff, Jeffrey M and Ivanov, Plamen Ch and Mark, Roger G and Mietus, Joseph E and Moody, George B and Peng, Chung-Kang and Stanley, H Eugene},
  journal={circulation},
  volume={101},
  number={23},
  pages={e215--e220},
  year={2000},
  publisher={Am Heart Assoc}
}

@article{dau2019ucr,
  title={The UCR time series archive},
  author={Dau, Hoang Anh and Bagnall, Anthony and Kamgar, Kaveh and Yeh, Chin-Chia Michael and Zhu, Yan and Gharghabi, Shaghayegh and Ratanamahatana, Chotirat Ann and Keogh, Eamonn},
  journal={IEEE/CAA Journal of Automatica Sinica},
  volume={6},
  number={6},
  pages={1293--1305},
  year={2019},
  publisher={IEEE}
}

@inproceedings{zhao2015classifying,
  title={Classifying phonological categories in imagined and articulated speech},
  author={Zhao, Shunan and Rudzicz, Frank},
  booktitle={2015 IEEE International Conference on Acoustics, Speech and Signal Processing (ICASSP)},
  pages={992--996},
  year={2015},
  organization={IEEE}
}

@article{nguyen2017inferring,
  title={Inferring imagined speech using EEG signals: a new approach using Riemannian manifold features},
  author={Nguyen, Chuong H and Karavas, George K and Artemiadis, Panagiotis},
  journal={Journal of neural engineering},
  volume={15},
  number={1},
  pages={016002},
  year={2017},
  publisher={IOP Publishing}
}

@article{johnson2016mimic,
  title={MIMIC-III, a freely accessible critical care database},
  author={Johnson, Alistair EW and Pollard, Tom J and Shen, Lu and Lehman, Li-wei H and Feng, Mengling and Ghassemi, Mohammad and Moody, Benjamin and Szolovits, Peter and Anthony Celi, Leo and Mark, Roger G},
  journal={Scientific data},
  volume={3},
  number={1},
  pages={1--9},
  year={2016},
  publisher={Nature Publishing Group}
}

@article{sharma2019dataset,
  title={A dataset of continuous affect annotations and physiological signals for emotion analysis},
  author={Sharma, Karan and Castellini, Claudio and van den Broek, Egon L and Albu-Schaeffer, Alin and Schwenker, Friedhelm},
  journal={Scientific data},
  volume={6},
  number={1},
  pages={196},
  year={2019},
  publisher={Nature Publishing Group UK London}
}

@inproceedings{zhang2020rcea,
  title={Rcea: Real-time, continuous emotion annotation for collecting precise mobile video ground truth labels},
  author={Zhang, Tianyi and El Ali, Abdallah and Wang, Chen and Hanjalic, Alan and Cesar, Pablo},
  booktitle={Proceedings of the 2020 CHI Conference on Human Factors in Computing Systems},
  pages={1--15},
  year={2020}
}

@article{xue2021ceap,
  title={CEAP-360VR: A Continuous Physiological and Behavioral Emotion Annotation Dataset for 360 VR Videos},
  author={Xue, Tong and El Ali, Abdallah and Zhang, Tianyi and Ding, Gangyi and Cesar, Pablo},
  journal={IEEE Transactions on Multimedia},
  year={2021},
  publisher={IEEE}
}

@phdthesis{shoeb2009application,
  title={Application of machine learning to epileptic seizure onset detection and treatment},
  author={Shoeb, Ali Hossam},
  year={2009},
  school={Massachusetts Institute of Technology}
}

@article{petrutiu2007abrupt,
  title={Abrupt changes in fibrillatory wave characteristics at the termination of paroxysmal atrial fibrillation in humans},
  author={Petrutiu, Simona and Sahakian, Alan V and Swiryn, Steven},
  journal={Europace},
  volume={9},
  number={7},
  pages={466--470},
  year={2007},
  publisher={Oxford University Press}
}

@article{brown2020language,
  title={Language models are few-shot learners},
  author={Brown, Tom and Mann, Benjamin and Ryder, Nick and Subbiah, Melanie and Kaplan, Jared D and Dhariwal, Prafulla and Neelakantan, Arvind and Shyam, Pranav and Sastry, Girish and Askell, Amanda and others},
  journal={Advances in neural information processing systems},
  volume={33},
  pages={1877--1901},
  year={2020}
}

@article{jin2023time,
  title={Time-llm: Time series forecasting by reprogramming large language models},
  author={Jin, Ming and Wang, Shiyu and Ma, Lintao and Chu, Zhixuan and Zhang, James Y and Shi, Xiaoming and Chen, Pin-Yu and Liang, Yuxuan and Li, Yuan-Fang and Pan, Shirui and others},
  journal={arXiv preprint arXiv:2310.01728},
  year={2023}
}

@inproceedings{sendera2023hypershot,
  title={Hypershot: Few-shot learning by kernel hypernetworks},
  author={Sendera, Marcin and Przewi{\k{e}}{\'z}likowski, Marcin and Karanowski, Konrad and Zi{\k{e}}ba, Maciej and Tabor, Jacek and Spurek, Przemys{\l}aw},
  booktitle={Proceedings of the IEEE/CVF Winter Conference on Applications of Computer Vision},
  pages={2469--2478},
  year={2023}
}

@article{requeima2019fast,
  title={Fast and flexible multi-task classification using conditional neural adaptive processes},
  author={Requeima, James and Gordon, Jonathan and Bronskill, John and Nowozin, Sebastian and Turner, Richard E},
  journal={Advances in Neural Information Processing Systems},
  volume={32},
  year={2019}
}

@article{tian2024survey,
  title={A survey on few-shot class-incremental learning},
  author={Tian, Songsong and Li, Lusi and Li, Weijun and Ran, Hang and Ning, Xin and Tiwari, Prayag},
  journal={Neural Networks},
  volume={169},
  pages={307--324},
  year={2024},
  publisher={Elsevier}
}

@inproceedings{zhang2023prompt,
  title={Prompt, generate, then cache: Cascade of foundation models makes strong few-shot learners},
  author={Zhang, Renrui and Hu, Xiangfei and Li, Bohao and Huang, Siyuan and Deng, Hanqiu and Qiao, Yu and Gao, Peng and Li, Hongsheng},
  booktitle={Proceedings of the IEEE/CVF Conference on Computer Vision and Pattern Recognition},
  pages={15211--15222},
  year={2023}
}

@article{koelstra2011deap,
  title={Deap: A database for emotion analysis; using physiological signals},
  author={Koelstra, Sander and Muhl, Christian and Soleymani, Mohammad and Lee, Jong-Seok and Yazdani, Ashkan and Ebrahimi, Touradj and Pun, Thierry and Nijholt, Anton and Patras, Ioannis},
  journal={IEEE transactions on affective computing},
  volume={3},
  number={1},
  pages={18--31},
  year={2011},
  publisher={IEEE}
}

@article{wagenaar2015collaborating,
  title={Collaborating and sharing data in epilepsy research},
  author={Wagenaar, Joost B and Worrell, Gregory A and Ives, Zachary and Matthias, Duempelmann and Litt, Brian and Schulze-Bonhage, Andreas},
  journal={Journal of Clinical Neurophysiology},
  volume={32},
  number={3},
  pages={235--239},
  year={2015},
  publisher={LWW}
}

@inproceedings{murphy2009eeg,
  title={EEG responds to conceptual stimuli and corpus semantics},
  author={Murphy, Brian and Baroni, Marco and Poesio, Massimo},
  booktitle={Proceedings of the 2009 Conference on Empirical Methods in Natural Language Processing},
  pages={619--627},
  year={2009}
}

@article{kaneshiro2015representational,
  title={A representational similarity analysis of the dynamics of object processing using single-trial EEG classification},
  author={Kaneshiro, Blair and Perreau Guimaraes, Marcos and Kim, Hyung-Suk and Norcia, Anthony M and Suppes, Patrick},
  journal={Plos one},
  volume={10},
  number={8},
  pages={e0135697},
  year={2015},
  publisher={Public Library of Science San Francisco, CA USA}
}

@article{liu2021comparing, title={Comparing Recognition Performance and Robustness of Multimodal Deep Learning Models for Multimodal Emotion Recognition}, author={Liu, Wei and Qiu, Jie-Lin and Zheng, Wei-Long and Lu, Bao-Liang}, journal={IEEE Transactions on Cognitive and Developmental Systems}, year={2021}, publisher={IEEE} }

@inproceedings{song2016multimodal,
  title={Multimodal multi-stream deep learning for egocentric activity recognition},
  author={Song, Sibo and Chandrasekhar, Vijay and Mandal, Bappaditya and Li, Liyuan and Lim, Joo-Hwee and Sateesh Babu, Giduthuri and Phyo San, Phyo and Cheung, Ngai-Man},
  booktitle={Proceedings of the IEEE conference on computer vision and pattern recognition workshops},
  pages={24--31},
  year={2016}
}

@inproceedings{nakamura2017jointly,
  title={Jointly learning energy expenditures and activities using egocentric multimodal signals},
  author={Nakamura, Katsuyuki and Yeung, Serena and Alahi, Alexandre and Fei-Fei, Li},
  booktitle={Proceedings of the IEEE Conference on Computer Vision and Pattern Recognition},
  pages={1868--1877},
  year={2017}
}

@misc{liu2021csl,
  title={CSL-SHARE: A multimodal wearable sensor-based human activity dataset},
  author={Liu, Hui and Hartmann, Yale and Schultz, Tanja},
  year={2021},
  publisher={Frontiers Media SA}
}

@article{micucci2017unimib,
  title={Unimib shar: A dataset for human activity recognition using acceleration data from smartphones},
  author={Micucci, Daniela and Mobilio, Marco and Napoletano, Paolo},
  journal={Applied Sciences},
  volume={7},
  number={10},
  pages={1101},
  year={2017},
  publisher={MDPI}
}

@article{pizzolato2017comparison,
  title={Comparison of six electromyography acquisition setups on hand movement classification tasks},
  author={Pizzolato, Stefano and Tagliapietra, Luca and Cognolato, Matteo and Reggiani, Monica and M{\"u}ller, Henning and Atzori, Manfredo},
  journal={PloS one},
  volume={12},
  number={10},
  pages={e0186132},
  year={2017},
  publisher={Public Library of Science San Francisco, CA USA}
}

@article{tangermann2012review,
  title={Review of the BCI competition IV},
  author={Tangermann, Michael and M{\"u}ller, Klaus-Robert and Aertsen, Ad and Birbaumer, Niels and Braun, Christoph and Brunner, Clemens and Leeb, Robert and Mehring, Carsten and Miller, Kai J and Mueller-Putz, Gernot and others},
  journal={Frontiers in neuroscience},
  pages={55},
  year={2012},
  publisher={Frontiers}
}

@article{cho2017eeg,
  title={EEG datasets for motor imagery brain--computer interface},
  author={Cho, Hohyun and Ahn, Minkyu and Ahn, Sangtae and Kwon, Moonyoung and Jun, Sung Chan},
  journal={GigaScience},
  volume={6},
  number={7},
  pages={gix034},
  year={2017},
  publisher={Oxford University Press}
}

@article{schalk2004bci2000,
  title={BCI2000: a general-purpose brain-computer interface (BCI) system},
  author={Schalk, Gerwin and McFarland, Dennis J and Hinterberger, Thilo and Birbaumer, Niels and Wolpaw, Jonathan R},
  journal={IEEE Transactions on biomedical engineering},
  volume={51},
  number={6},
  pages={1034--1043},
  year={2004},
  publisher={IEEE}
}

@article{srivastava2018combining,
  title={Combining low and mid-level gaze features for desktop activity recognition},
  author={Srivastava, Namrata and Newn, Joshua and Velloso, Eduardo},
  journal={Proceedings of the ACM on Interactive, Mobile, Wearable and Ubiquitous Technologies},
  volume={2},
  number={4},
  pages={1--27},
  year={2018},
  publisher={ACM New York, NY, USA}
}

@inproceedings{kunze2013know,
  title={I know what you are reading: recognition of document types using mobile eye tracking},
  author={Kunze, Kai and Utsumi, Yuzuko and Shiga, Yuki and Kise, Koichi and Bulling, Andreas},
  booktitle={Proceedings of the 2013 international symposium on wearable computers},
  pages={113--116},
  year={2013}
}

@article{mattout2015bci,
  title={BCI Challenge},
  author={Mattout, J and Manu and Maucle and Kan, W},
  journal={Kaggle},
  year={2015},
  publisher={Kaggle}
}

@article{arico2014influence,
  title={Influence of P300 latency jitter on event related potential-based brain--computer interface performance},
  author={Aric{\`o}, P and Aloise, F and Schettini, F and Salinari, S and Mattia, D and Cincotti, F},
  journal={Journal of neural engineering},
  volume={11},
  number={3},
  pages={035008},
  year={2014},
  publisher={IOP Publishing}
}

@inproceedings{duan2013differential,
  title={Differential entropy feature for EEG-based emotion classification},
  author={Duan, Ruo-Nan and Zhu, Jia-Yi and Lu, Bao-Liang},
  booktitle={2013 6th International IEEE/EMBS Conference on Neural Engineering (NER)},
  pages={81--84},
  year={2013},
  organization={IEEE}
}

@article{andrzejak2001indications,
  title={Indications of nonlinear deterministic and finite-dimensional structures in time series of brain electrical activity: Dependence on recording region and brain state},
  author={Andrzejak, Ralph G and Lehnertz, Klaus and Mormann, Florian and Rieke, Christoph and David, Peter and Elger, Christian E},
  journal={Physical Review E},
  volume={64},
  number={6},
  pages={061907},
  year={2001},
  publisher={APS}
}

@misc{shoeb2000chb,
  title={Chb-mit scalp eeg database},
  author={Shoeb, A},
  year={2000},
  publisher={MIT. https://www. physionet. org/pn6/chbmit}
}

@inproceedings{clifford2017af,
  title={AF classification from a short single lead ECG recording: The PhysioNet/computing in cardiology challenge 2017},
  author={Clifford, Gari D and Liu, Chengyu and Moody, Benjamin and Li-wei, H Lehman and Silva, Ikaro and Li, Qiao and Johnson, AE and Mark, Roger G},
  booktitle={2017 Computing in Cardiology (CinC)},
  pages={1--4},
  year={2017},
  organization={IEEE}
}

@article{bousseljot1995nutzung,
  title={Nutzung der EKG-Signaldatenbank CARDIODAT der PTB {\"u}ber das Internet},
  author={Bousseljot, Ralf and Kreiseler, Dieter and Schnabel, Allard},
  year={1995},
  publisher={Walter de Gruyter, Berlin/New York Berlin, New York}
}

@article{pollard2018eicu,
  title={The eICU Collaborative Research Database, a freely available multi-center database for critical care research},
  author={Pollard, Tom J and Johnson, Alistair EW and Raffa, Jesse D and Celi, Leo A and Mark, Roger G and Badawi, Omar},
  journal={Scientific data},
  volume={5},
  number={1},
  pages={1--13},
  year={2018},
  publisher={Nature Publishing Group}
}

@article{khan2021ecg,
  title={ECG Images dataset of Cardiac and COVID-19 Patients},
  author={Khan, Ali Haider and Hussain, Muzammil and Malik, Muhammad Kamran},
  journal={Data in Brief},
  volume={34},
  pages={106762},
  year={2021},
  publisher={Elsevier}
}

@article{schaette2011tinnitus,
  title={Tinnitus with a normal audiogram: physiological evidence for hidden hearing loss and computational model},
  author={Schaette, Roland and McAlpine, David},
  journal={Journal of Neuroscience},
  volume={31},
  number={38},
  pages={13452--13457},
  year={2011},
  publisher={Soc Neuroscience}
}

@article{guest2017tinnitus,
  title={Tinnitus with a normal audiogram: Relation to noise exposure but no evidence for cochlear synaptopathy},
  author={Guest, Hannah and Munro, Kevin J and Prendergast, Garreth and Howe, Simon and Plack, Christopher J},
  journal={Hearing research},
  volume={344},
  pages={265--274},
  year={2017},
  publisher={Elsevier}
}

@article{sun2023few,
  title={Few-Shot Class-Incremental Learning for Medical Time Series Classification},
  author={Sun, Le and Zhang, Mingyang and Wang, Benyou and Tiwari, Prayag},
  journal={IEEE Journal of Biomedical and Health Informatics},
  year={2023},
  publisher={IEEE}
}

@inproceedings{moin2018emg,
  title={An EMG gesture recognition system with flexible high-density sensors and brain-inspired high-dimensional classifier},
  author={Moin, Ali and Zhou, Andy and Rahimi, Abbas and Benatti, Simone and Menon, Alisha and Tamakloe, Senam and Ting, Jonathan and Yamamoto, Natasha and Khan, Yasser and Burghardt, Fred and others},
  booktitle={2018 IEEE International Symposium on Circuits and Systems (ISCAS)},
  pages={1--5},
  year={2018},
  organization={IEEE}
}

@article{hoffmann2008efficient,
  title={An efficient P300-based brain--computer interface for disabled subjects},
  author={Hoffmann, Ulrich and Vesin, Jean-Marc and Ebrahimi, Touradj and Diserens, Karin},
  journal={Journal of Neuroscience methods},
  volume={167},
  number={1},
  pages={115--125},
  year={2008},
  publisher={Elsevier}
}

@article{arjovsky2017towards,
  title={Towards principled methods for training generative adversarial networks},
  author={Arjovsky, Martin and Bottou, L{\'e}on},
  journal={arXiv preprint arXiv:1701.04862},
  year={2017}
}

@inproceedings{hinz2021improved,
  title={Improved techniques for training single-image gans},
  author={Hinz, Tobias and Fisher, Matthew and Wang, Oliver and Wermter, Stefan},
  booktitle={Proceedings of the IEEE/CVF Winter Conference on Applications of Computer Vision},
  pages={1300--1309},
  year={2021}
}

@article{gruver2023large,
  title={Large language models are zero-shot time series forecasters},
  author={Gruver, Nate and Finzi, Marc and Qiu, Shikai and Wilson, Andrew Gordon},
  journal={arXiv preprint arXiv:2310.07820},
  year={2023}
}

@article{hou2022metaprompting,
  title={MetaPrompting: Learning to learn better prompts},
  author={Hou, Yutai and Dong, Hongyuan and Wang, Xinghao and Li, Bohan and Che, Wanxiang},
  journal={arXiv preprint arXiv:2209.11486},
  year={2022}
}

@article{sempionatto2021epidermal,
  title={An epidermal patch for the simultaneous monitoring of haemodynamic and metabolic biomarkers},
  author={Sempionatto, Juliane R and Lin, Muyang and Yin, Lu and De la Paz, Ernesto and Pei, Kexin and Sonsa-Ard, Thitaporn and de Loyola Silva, Andre N and Khorshed, Ahmed A and Zhang, Fangyu and Tostado, Nicholas and others},
  journal={Nature Biomedical Engineering},
  volume={5},
  number={7},
  pages={737--748},
  year={2021},
  publisher={Nature Publishing Group UK London}
}

@article{xu2024physicochemical,
  title={A physicochemical-sensing electronic skin for stress response monitoring},
  author={Xu, Changhao and Song, Yu and Sempionatto, Juliane R and Solomon, Samuel A and Yu, You and Nyein, Hnin YY and Tay, Roland Yingjie and Li, Jiahong and Heng, Wenzheng and Min, Jihong and others},
  journal={Nature Electronics},
  pages={1--12},
  year={2024},
  publisher={Nature Publishing Group UK London}
}

@inproceedings{radford2021learning,
  title={Learning transferable visual models from natural language supervision},
  author={Radford, Alec and Kim, Jong Wook and Hallacy, Chris and Ramesh, Aditya and Goh, Gabriel and Agarwal, Sandhini and Sastry, Girish and Askell, Amanda and Mishkin, Pamela and Clark, Jack and others},
  booktitle={International conference on machine learning},
  pages={8748--8763},
  year={2021},
  organization={PMLR}
}

@inproceedings{caron2021emerging,
  title={Emerging properties in self-supervised vision transformers},
  author={Caron, Mathilde and Touvron, Hugo and Misra, Ishan and J{\'e}gou, Herv{\'e} and Mairal, Julien and Bojanowski, Piotr and Joulin, Armand},
  booktitle={Proceedings of the IEEE/CVF international conference on computer vision},
  pages={9650--9660},
  year={2021}
}

@inproceedings{ramesh2021zero,
  title={Zero-shot text-to-image generation},
  author={Ramesh, Aditya and Pavlov, Mikhail and Goh, Gabriel and Gray, Scott and Voss, Chelsea and Radford, Alec and Chen, Mark and Sutskever, Ilya},
  booktitle={International Conference on Machine Learning},
  pages={8821--8831},
  year={2021},
  organization={PMLR}
}

@article{simonyan2014very,
  title={Very deep convolutional networks for large-scale image recognition},
  author={Simonyan, Karen and Zisserman, Andrew},
  journal={arXiv preprint arXiv:1409.1556},
  year={2014}
}

@inproceedings{huang2017densely,
  title={Densely connected convolutional networks},
  author={Huang, Gao and Liu, Zhuang and Van Der Maaten, Laurens and Weinberger, Kilian Q},
  booktitle={Proceedings of the IEEE conference on computer vision and pattern recognition},
  pages={4700--4708},
  year={2017}
}

@inproceedings{he2015delving,
  title={Delving deep into rectifiers: Surpassing human-level performance on imagenet classification},
  author={He, Kaiming and Zhang, Xiangyu and Ren, Shaoqing and Sun, Jian},
  booktitle={Proceedings of the IEEE international conference on computer vision},
  pages={1026--1034},
  year={2015}
}

@inproceedings{tan2019efficientnet,
  title={Efficientnet: Rethinking model scaling for convolutional neural networks},
  author={Tan, Mingxing and Le, Quoc},
  booktitle={International conference on machine learning},
  pages={6105--6114},
  year={2019},
  organization={PMLR}
}

@inproceedings{szegedy2016rethinking,
  title={Rethinking the inception architecture for computer vision},
  author={Szegedy, Christian and Vanhoucke, Vincent and Ioffe, Sergey and Shlens, Jon and Wojna, Zbigniew},
  booktitle={Proceedings of the IEEE conference on computer vision and pattern recognition},
  pages={2818--2826},
  year={2016}
}

@inproceedings{chen2020simple,
  title={A simple framework for contrastive learning of visual representations},
  author={Chen, Ting and Kornblith, Simon and Norouzi, Mohammad and Hinton, Geoffrey},
  booktitle={International conference on machine learning},
  pages={1597--1607},
  year={2020},
  organization={PMLR}
}

@article{baevski2020wav2vec,
  title={wav2vec 2.0: A framework for self-supervised learning of speech representations},
  author={Baevski, Alexei and Zhou, Yuhao and Mohamed, Abdelrahman and Auli, Michael},
  journal={Advances in neural information processing systems},
  volume={33},
  pages={12449--12460},
  year={2020}
}

@article{devlin2018bert,
  title={Bert: Pre-training of deep bidirectional transformers for language understanding},
  author={Devlin, Jacob and Chang, Ming-Wei and Lee, Kenton and Toutanova, Kristina},
  journal={arXiv preprint arXiv:1810.04805},
  year={2018}
}

@article{joshi2020spanbert,
  title={Spanbert: Improving pre-training by representing and predicting spans},
  author={Joshi, Mandar and Chen, Danqi and Liu, Yinhan and Weld, Daniel S and Zettlemoyer, Luke and Levy, Omer},
  journal={Transactions of the association for computational linguistics},
  volume={8},
  pages={64--77},
  year={2020},
  publisher={MIT Press One Rogers Street, Cambridge, MA 02142-1209, USA journals-info~…}
}

@article{PhysioNet,
 author = "Goldberger, A. L. and Amaral, L. A. N. and Glass, L. and
	   Hausdorff, J. M. and Ivanov, P. Ch. and Mark, R. G. and
	   Mietus, J. E. and Moody, G. B. and Peng, C.-K. and Stanley, H. E.",
 title = "{PhysioBank, PhysioToolkit, and PhysioNet}: Components of a New
	  Research Resource for Complex Physiologic Signals",
 journal = "Circulation",
 year = "2000 (June 13)",
 volume = "101",
 number = "23",
 pages = "e215--e220",
 note = "Circulation Electronic Pages:
         http://circ.ahajournals.org/content/101/23/e215.full
         PMID:1085218; doi: 10.1161/01.CIR.101.23.e215"}

\end{document}